\documentclass[lettersize,journal]{IEEEtran}
\usepackage[numbers]{natbib}
\usepackage{multicol}
\usepackage{multirow}
\usepackage[bookmarks=true]{hyperref}
\usepackage{url}
\usepackage{booktabs}
\usepackage{amssymb}
\usepackage{amsmath}
\usepackage{graphicx}
\usepackage{subcaption}
\usepackage{siunitx}
\usepackage{placeins}
\usepackage{xcolor}  
\begin{document}
\bstctlcite{IEEEexample:BSTcontrol}

\title{FALCON-S: Fixed-wing ground-effect Aerodynamics Simulator and Flight Control Learning Suite}

\author{Matteo El Hariry\textsuperscript{1} \and Pedro Lima\textsuperscript{1} \and
        Andrej Orsula\textsuperscript{1} \and Matthieu Geist\textsuperscript{2} \and
        Miguel Olivares-Mendez\textsuperscript{1},
  \thanks{\textsuperscript{1} Interdisciplinary Centre for Security, Reliability and Trust (SnT),
    University of Luxembourg, L-1855 Luxembourg
    (e-mail: \{matteo.elhariry@uni.lu).}
  \thanks{\textsuperscript{2} Earth Speacies Project.}
  \thanks{This work was supported by the European Union under Grant Agreement No.\ 101096487.}
  \thanks{Code will be released upon publication.}
}


\maketitle

\begin{abstract}
We present a modular, high-fidelity simulation framework for the development and benchmarking of flight control strategies in fixed-wing aerial robots operating near the ground. Unlike existing simulators that rely on simplified or hover-oriented dynamics, our framework models full 6DoF rigid-body physics, semi-empirical ground-effect aerodynamics, actuator dynamics, sensor noise, and environmental disturbances. This physical realism, combined with modular component design, enables systematic analysis of low-altitude flight behavior under realistic conditions. The simulator supports both CPU and GPU backends via Torch and NVIDIA Warp, enabling high-throughput parallel execution suitable for large-scale reinforcement learning training and optimal control rollouts. A unified interface accommodates a range of controllers (both RL and optical control algorithms) across tasks such as altitude regulation and trajectory tracking. Cross-validation with X-Plane and JSBSim is also supported to facilitate engineering integration and visual fidelity. 
\end{abstract}

\vspace{-0.3cm}


\section{Introduction}
The design of autonomous control systems for aerial vehicles has increasingly relied on simulation as a primary development and evaluation tool. Across robotics, advances in optimal control, sampling-based methods, and reinforcement learning (RL) have enabled increasingly capable controllers for complex platforms, including quadrotors~\cite{kaufmann2023champion}, fixed-wing UAVs~\cite{bohn2019deep, de2023deep}, satellites~\cite{el2024drift}, and legged robots~\cite{chane2024cat}. Despite this progress, deploying autonomous controllers on aerial vehicles, particularly in low-altitude and near-ground regimes, remains challenging due to discrepancies between simplified simulation models and the physical effects encountered in real flight, such as actuator dynamics, sensing imperfections, and aerodynamic interactions with the environment.
Existing simulation environments, including \textit{JSBSim}~\cite{berndt2004jsbsim}, \textit{X-Plane}~\cite{xplane}, and \textit{Flightmare}~\cite{song2021flightmare}, offer valuable capabilities but are typically optimized for specific use cases. High-fidelity simulators emphasize realism and pilot training but are difficult to scale for large-scale algorithmic evaluation, while learning-oriented simulators often prioritize throughput at the expense of aerodynamic fidelity and sub-system level realism. As a result, key phenomena relevant to aerial robotics, such as ground effect, realistic actuator, sensor behavior, and controlled variation of physical assumptions, are often difficult to study systematically within a unified framework. This limits the ability to benchmark autonomous flight controllers across different modeling assumptions, task complexities, and control paradigms.
\begin{figure}[t!]
    \centering
    \includegraphics[width=1.0\linewidth]{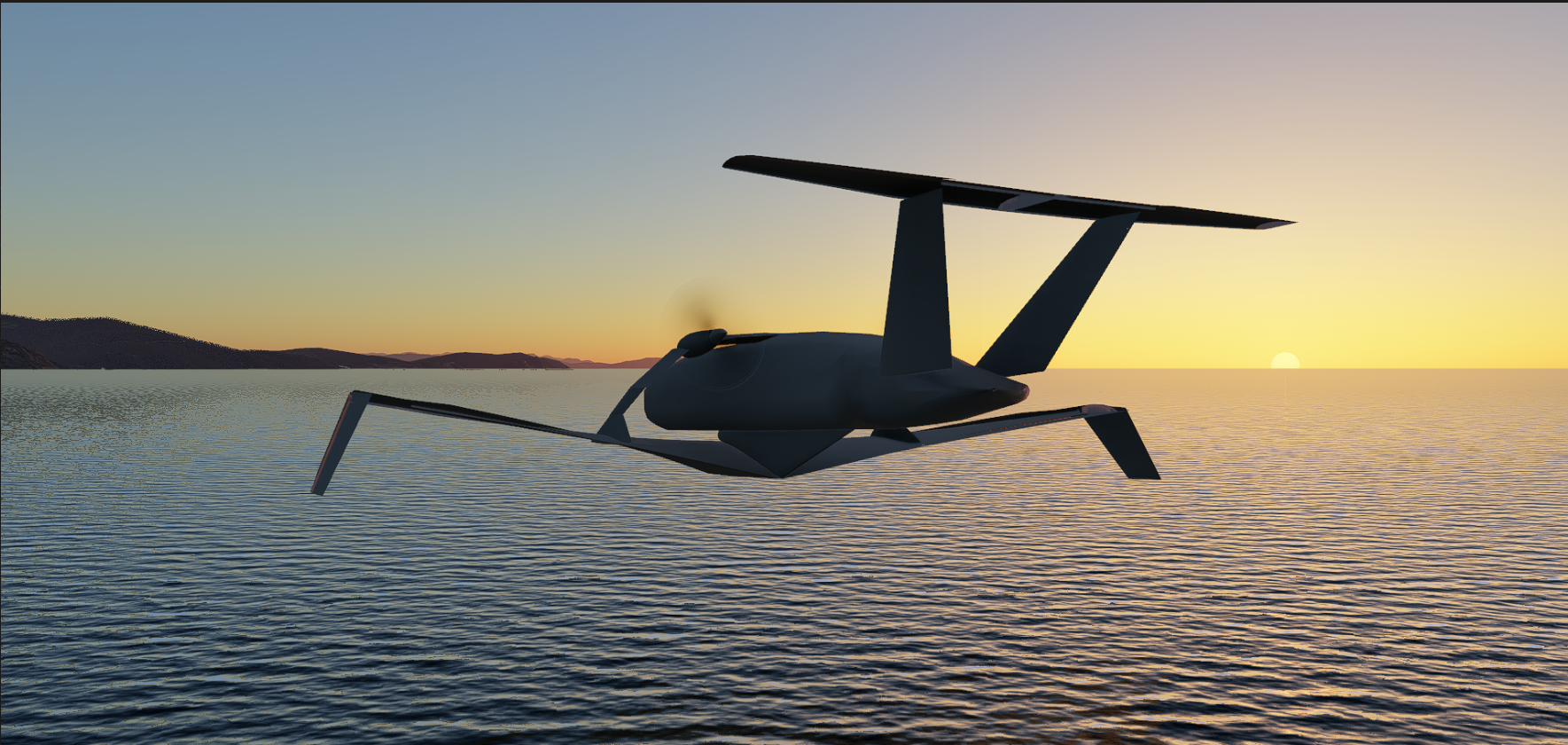}
    \caption{Airship flying in ground effect, rendered in X-Plane.}
    \label{fig:hero}
    \vspace{-0.7cm}
\end{figure}

In this work, we introduce a modular, physics-grounded simulation benchmark for fixed-wing aerial
vehicles, designed to support systematic evaluation of autonomous control methods under realistic
low-altitude flight conditions. The benchmark emphasizes ground-effect aerodynamics, actuator and
sensor realism, and full 6DoF dynamics, while remaining compatible with both model-based and
learning-based control approaches. Our simulator provides:\\
(1) \textbf{Scalable multi-backend simulation engine:} We implement a GPU-accelerated physics pipeline
using NVIDIA Warp \cite{warp2022}, alongside a pure-PyTorch implementation of the same dynamics
(validated to single-step numerical parity) and a CPU-compatible fallback, enabling high-throughput
simulation for both learning-based and classical control methods. 
The Warp backend sustains $19.5$ million environment-steps per second at $16{,}384$ parallel environments 
corresponding to a measured $13.4\times$ speed-up w.r.t.\ the current state-of-the-art~\cite{xue2024neuralplane}. \\
(2) \textbf{Modular architecture for control and benchmarking:} The framework supports both
reinforcement learning and optimal control methods, and includes a library of physically grounded
flight tasks (e.g., altitude regulation, trajectory tracking, attitude-command tracking) defined over
five identified airframes, that can be easily extended or modified.\\
(3) \textbf{Comprehensive physics modeling:} Our simulation core incorporates aerodynamic models
automatically pre-derived from the aircraft's geometry, and accounts for ground effect, wind, and
atmospheric density, along with configurable sensor and actuator dynamics to support high-fidelity
experiments.\\
(4) \textbf{Cross-platform validation interfaces:} We provide validation capabilities with
JSBSim, and X-Plane (for closed-loop testing with high-fidelity rendering), broadening applicability across academic and industrial settings.

\begin{table*}[t!]
\centering
\small
\caption{Comparison with existing aircraft simulation frameworks. Our platform combines near-ground aerodynamics, modular tasks, advanced controllers, rich sensor and actuator models. [\checkmark] full, [*] partial/optional, [--] not supported.}
\label{tab:sim_comparison}
\footnotesize
\setlength{\tabcolsep}{5pt}

\begin{tabular}{lcccccc}
\toprule
\textbf{Feature} & \textbf{Ours} & \textbf{NeuralPlane} & \textbf{QPlane} & \textbf{JSBSim} & \textbf{XPlane}\\
\midrule
\textbf{Open-source}         & \checkmark & \checkmark & \checkmark & \checkmark & -- \\
\textbf{Physics-based FDM}   & \checkmark (WIG, 6DoF) & \checkmark (fixed-wing only) & \checkmark (JSBSim/XPlane) & \checkmark & \checkmark \\
\textbf{Ground Effect Model} & \checkmark (semi-empirical) & -- & * (depends on JSBSim) & * & \checkmark \\
\textbf{GPU Acceleration}    & \checkmark (Warp) & \checkmark (PyTorch) & -- & -- & -- \\
\textbf{Multi-agent Support} & -- & \checkmark & \checkmark & * & * (via UDP) \\
\textbf{Multiple Flight Tasks}        & \checkmark  & \checkmark & \checkmark & * & * \\
\textbf{Controller Support}  & \checkmark & \checkmark & \checkmark & * & * \\
\textbf{Realism}             & High & Medium & High (JSBSim/XPlane) & High & High \\
\textbf{Visualization Tools} & \checkmark & * (via FlightGear/Tacview) & * (via XPlane/FlightGear) & * (via FlightGear) & \checkmark \\
\textbf{Sim-to-Real Enabling}   & * & * & * & \checkmark & \checkmark \\
\bottomrule
\end{tabular}
\end{table*}


\section{Related Work}

\textbf{Simulation of fixed-wing flight dynamics.} 
Simulators such as \textit{JSBSim}~\cite{berndt2004jsbsim}, \textit{FlightGear}~\cite{perry2004flightgear}, and \textit{X-Plane}~\cite{xplane} have long supported fixed-wing aircraft modeling, but are primarily designed for pilot training or certification, and lack native support for reinforcement learning or scalable training. Recent research platforms such as QPlane~\cite{richter2021qplane} and NeuralPlane~\cite{xue2024neuralplane} address this limitation by exposing lightweight and configurable interfaces suitable for policy learning: QPlane wraps X-Plane and JSBSim for Gym-based RL experiments, while NeuralPlane introduces a parallel GPU-based pipeline for
efficient large-scale simulation. However, these frameworks either simplify the flight dynamics, omitting ground-effect modeling, sensor and actuator fidelity, and environmental realism, or, in the case of QPlane, cannot efficiently execute multiple simulation environments in parallel.\\
\textbf{Accelerated simulators and learning environments.}
GPU-accelerated simulators such as IsaacGym~\cite{makoviychuk2021isaac} and 
WarpDrive~\cite{pan2021warpdrive} have become central to robotics research, but few target flight
vehicles. Platforms like AirSim~\cite{airsim2017}, RotorS~\cite{furrer2016rotors}, and
Flightmare~\cite{song2021flightmare} (GPU-accelerated via Unity~\cite{juliani2018unity}) have advanced
learning-based control for multirotors, yet fixed-wing benchmarks remain scarce due to the complexity
of forward-flight dynamics, non-holonomic constraints, and sensitivity to external disturbances~\cite{kaufmann2023champion}. Our Warp-based simulator provides domain-specific GPU
acceleration for fixed-wing vehicles with detailed aerodynamics, supporting large-scale training
without compromising physical realism.\\
\textbf{Reinforcement learning for fixed-wing flight control.}
RL has been applied to fixed-wing attitude and altitude control with PPO~\cite{bohn2019deep} and
DDPG-family methods~\cite{demarco2023deep, shukla2025evolving, yao2026physical, khanzada2025altitude},
including recent sim-to-real demonstrations via dynamic randomization and control-rate reward
shaping~\cite{chowdhury2024interchangeable}. A recent survey of the field~\cite{richter2024review}
identifies two open problems directly motivating this work: the absence of standardization as
simulators, airframes, state--action spaces, and metrics differ across papers, making results 
incomparable, and the fact that most learned controllers never leave simulation. Moreover, existing
studies target conventional airframes in free air; none address the near-ground regime where ground
effect alters the vehicle dynamics. Our benchmark provides a common, physically grounded evaluation
setting for these methods, spanning multiple airframes, tasks, and both on- and off-policy
algorithms.\\
\textbf{Unified benchmarking of classical and learned control.}
There is increasing interest in combining optimal control with reinforcement
learning~\cite{berkenkamp2019safe, peng2018sim}, yet direct comparisons under identical dynamics remain
rare. Table~\ref{tab:sim_comparison} situates our simulator against widely used
frameworks~\cite{xue2024neuralplane, richter2021qplane, berndt2004jsbsim, xplane}: while these
platforms offer high-fidelity dynamics, industry-grade validation, or large-scale parallelism, key
phenomena for near-ground autonomous flight such as explicit ground-effect aerodynamics, realistic actuator
and sensor modeling, controlled disturbance injection, and systematic variation of physical
assumptions, are difficult to study within a single extensible framework. Our work provides a fully open benchmark combining 6DoF dynamics, semi-empirical ground-effect corrections, actuator and sensor dynamics, wind turbulence, atmospheric effects, and OpenVSP-derived~\cite{openvsp} aerodynamic coefficients, with scalable CPU/GPU execution supporting classical baselines (LQR, MPPI) and learning-based policies under consistent, configurable conditions.

\begin{figure*}[ht!]
    \centering
    \includegraphics[width=.9 \textwidth, height=.4\textwidth]{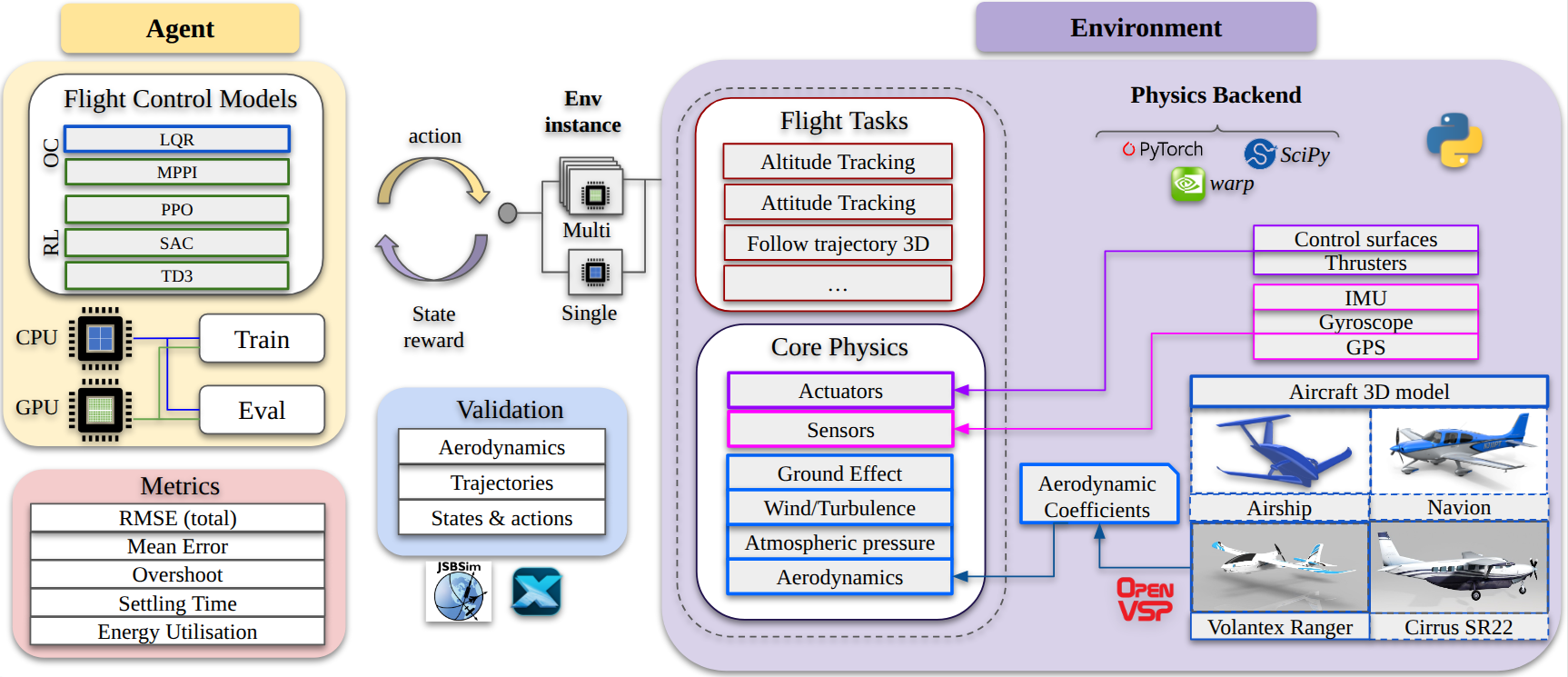}
    \small
\caption{\small Ground-effect fixed-wing simulation architecture. The modular environment models aerodynamics, actuators, ground effect, turbulence, and sensors, while the agent module supports classical and learning-based controllers. Tasks, metrics, and visualization enable extensible benchmarking and training.}
    \label{fig:system_architecture}
    \vspace{-.5cm}
\end{figure*}


\section{Preliminaries}\label{sec:preliminaries}
We consider the control of a rigid fixed-wing vehicle flying in proximity to the ground, modeled as a six-degrees-of-freedom (6DoF) system with coupled translational and rotational dynamics. The vehicle is subject to forces from gravity, aerodynamics, and propulsion, and its motion is described in the body frame.
The state vector $\mathbf{x} \in \mathbb{R}^{9} \times  \mathit{S}^{3}$ (or $\mathbf{x} \in \mathbb{R}^{12}$) comprises the position $\mathbf{p} \in \mathbb{R}^3$, orientation (represented as a unit quaternion $\mathbf{q} \in \mathit{S}^{3} = \left\{ \mathbf{q} \in \mathbb{H}: ||\mathbf{q}|| = 1 \right\}$ or Euler angles $\left( \phi, \theta, \psi \right) \in \mathbb{R}^3$), linear velocity $\mathbf{v} \in \mathbb{R}^3$, and angular velocity $\boldsymbol{\omega} \in \mathbb{R}^3$. Control inputs include throttle and actuator deflections for the elevator, rudder, and ailerons. The equations of motion follow Newton-Euler rigid body dynamics:
\begin{align}
    m \dot{\mathbf{v}} &= \mathbf{F}_g + \mathbf{F}_a + \mathbf{F}_t - \boldsymbol{\omega} \times m \mathbf{v}, \\
    \mathbf{J} \dot{\boldsymbol{\omega}} &= \boldsymbol{\tau}_a + \boldsymbol{\tau}_t - \boldsymbol{\omega} \times \mathbf{J} \boldsymbol{\omega},
\end{align}
where $m$ is the vehicle mass,  $\mathbf{J}$ is the inertia tensor, $\mathbf{F}_g$ is the gravitational force, $\mathbf{F}_a$ and $\boldsymbol{\tau}_a$ are aerodynamic forces and moments, and $\mathbf{F}_t$ and $\boldsymbol{\tau}_t$ are thrust-generated force and moment vectors. We assume constant mass and neglect gyroscopic effects. Each vehicle is modeled as a rigid body with a body-fixed frame $\left\{b\right\}$ rigidly attached at the centre of mass, and motion is described relative to an inertial north--east--down (NED) frame $\left\{I\right\}$.

\subsection{Aerodynamic Model}

Aerodynamic forces and moments are computed using semi-empirical models based on the vehicle's angle of attack $\alpha$, sideslip $\beta$, Reynolds number $Re$ and control surface deflections. The forces and moments take the form:
\begin{equation}
    \mathbf{F}_\text{aero} = q S \begin{bmatrix} -C_D \\ C_Y \\ -C_L \end{bmatrix}, \quad
    \mathbf{M}_\text{aero} = q S \begin{bmatrix} b C_l \\ c C_m \\ b C_n \end{bmatrix},
\end{equation}
where $q = \frac{1}{2} \rho V_a^2$ is the dynamic pressure, $S$ is the reference wing area, $b$ and $c$ are the wingspan and chord, and $C_i$ are the aerodynamic coefficients dependent on angle of attack $\alpha$, sideslip $\beta$, and control surfaces $\delta_a$ (ailerons), $\delta_e$ (elevator) and $\delta_r$ (rudder). Lift, drag, and side force coefficients are computed from look-up tables or parametric expressions derived from geometric tools such as OpenVSP~\cite{openvsp}.

\subsection{Ground Effect Model}\label{subsection:groundeffect}

Our simulator models ground effect, which alters the lift and drag characteristics of the vehicle when flying close to the surface, through the semi-empirical lifting-line corrections
of~\cite{phillips2013lifting}:
\begin{align}
  C_L &= C_L^{\infty} \left(1 + \mu_{L}(h/b)\right), \\
  C_D &= C_D^{\infty} \left(1 - \mu_{D}(h/b)\right)\left(1 + \mu_{L}(h/b)\right)^2,
\end{align}
where $C_L^\infty$, $C_D^\infty$ denote the out-of-ground-effect coefficients and $\mu_{L}, \mu_{D}$ are modifiers parameterized by the height-over-span ratio $h/b$, taper ratio, and aspect ratio (full parameterization in the open-source repository). The closed-form structure is directly compatible with GPU kernel execution (no panel-method or CFD computation) while capturing the key physical trends of increased lift and reduced induced drag in ground proximity. The correction can be toggled per environment, enabling the controlled ground-effect ablations of Section~\ref{sec:ge-energy}.

Overall, the simulator produces time-continuous dynamics discretized via a configurable integration scheme (e.g., Euler or RK4), exposed through a modular interface with both CPU and GPU implementations, and forming the basis for the environments used to train classical and learning-based controllers.


\section{Ground-Effect Fixed-Wing Flight Simulation Framework}

Our simulation platform, as illustrated in Fig.~\ref{fig:system_architecture}, consists of two primary modules: the \emph{agent} and the \emph{environment}. The first module hosts the controllers together with the training and evaluation pipelines that execute them on CPU or GPU. The environment module integrates a high-fidelity 6DoF flight model with realistic actuator dynamics, sensor imperfections, wind turbulence (Dryden), and optional ground effect modeling, exposed through three interchangeable physics backends (SciPy, PyTorch, and NVIDIA Warp). Both modules are configured via a modular system in which each physical phenomenon can be independently toggled, supporting controlled ablation studies across tasks and airframes.

\subsection{Agent Module}
The agent module supports different control models, including classical approaches such as Linear Quadratic Regulator (LQR) and Model Predictive Path Integral (MPPI)~\cite{mppi_paper}, as well as modern learning-based control policies derived from methods such as PPO~\cite{Schulman2017PPO}, SAC~\cite{haarnoja2018soft}, and TD3~\cite{fujimoto2018addressing-td3}. These controllers can be executed on either CPU or GPU for both evaluation and large-scale training, enabling both quick debugging of simulated flight conditions and heavy-duty batched experiments, where millions of trajectories can be collected to train and evaluate control performance.

\textbf{LQR with Integral action}:
The LQR controller is implemented in closed form using discrete-time linearization of the vehicle dynamics around a steady trim condition. The gain matrix $K$ is precomputed using the Riccati equation solution, and the resulting control law $u = -Kx$ is applied at each simulation step. The linearization matrices $(A, B)$ are precomputed and approximated using numerical Jacobians based on the simulator's physics model. For LQRI (LQR with integral action) the state vector is augmented with integrator states (e.g., integrated altitude error) before forming $(A_d, B_d)$ and solving the Riccati equation.

The controller was tuned using state and input weighting matrices selected to balance tracking accuracy and control effort. The state weighting matrix $Q_{LQR}$, the complete state-feedback gain matrix and the way they have been devised and tuned are provided in details on the open-sourced repository.
  Since the approach requires a stabilizable linearization about trim, its applicability is
  airframe-dependent (see Section~\ref{sec:experiments}).


\textbf{MPPI}:
The Model Predictive Path Integral controller follows a sampling-based trajectory optimization approach. At each control step, the algorithm samples multiple control sequences from a Gaussian distribution centred on the previous action sequence, propagates each action through the dynamics model, and computes an optimal control output from the weighted average of trajectories based on their cumulative cost. The implementation supports GPU-based sampling for parallelized inference, using a Warp backend. The cost function is task-specific and includes weighted penalties on tracking error, control effort, and constraint violations.

\textbf{Gymnasium interface}:  
The environment exposes a compliant Gymnasium~\cite{towers2024gymnasium} interface through the \texttt{CoreAircraftEnv} class and its wrappers. It supports \texttt{reset()}, \texttt{step(action)}, and \texttt{render()} methods, and optionally includes \texttt{info} dictionaries with task-specific diagnostics. Observations are exposed as flat NumPy arrays and can be extended with sensor noise or delays via wrapper classes. The action space is continuous (bounded) and directly maps to control surface deflections and throttle values.
For high-performance training and evaluation, parallelized variants of the environment are available
  through the Warp and Torch backends. These wrappers implement the same Gym API but execute
  dynamics in batched form on GPU, leveraging kernel-level integration and the GPU memory model.

 \textbf{RL algorithms}:
  We provide vectorized re-implementations of PPO, SAC and TD3 in the style of 
  CleanRL~\cite{huang2022cleanrl}, operating directly on the batched GPU environments; the
  Gymnasium-compliant CPU environment additionally remains compatible with off-the-shelf libraries such
  as Stable-Baselines3. GPU acceleration is used for both training and evaluation.


\subsection{Environment Module}
Our environment module supports multiple simulation backends and interoperation with external tools,
allowing flexibility in simulation fidelity, performance, and controller design workflows.
Specifically, we offer three physics backends in Python: one based on SciPy's numerical integration for easier prototyping, a pure-PyTorch implementation of the same dynamics for GPU execution within
standard deep-learning tooling, and one leveraging Warp~\cite{warp2022} for large-scale
GPU-accelerated simulation. The PyTorch and Warp backends are validated to single-step numerical
parity. 
This multi-backend setup enables practitioners to prototype quickly in Python and validate or deploy controllers using industry-standard tools when necessary.

\textbf{Core Physics}:
The environment simulates full six-degree-of-freedom (6DoF) rigid-body aircraft dynamics, focusing on low-altitude scenarios where physical effects such as ground proximity and turbulence dominate. The physics module is structured around five interconnected components:

\textit{Aerodynamics}: Uses precomputed aerodynamic coefficients from OpenVSP. Aerodynamic forces and moments are adjusted dynamically based on airspeed, angle of attack, sideslip angle, height above ground and control surface deflection. Ground effect corrections are applied using the semi-empirical model described in Section~\ref{subsection:groundeffect}.

\textit{Actuators}: Modular design allows the definition of an arbitrary number and type of
motors and control surfaces. Actuator systems are modeled as first- or second-order
dynamics with configurable time constants and damping ratios, so that latency, saturation, and
bandwidth-limited slew behaviour are reproduced:
\begin{equation}
  H(s) = \frac{\omega_n^2}{s^2 + 2\zeta\omega_n s + \omega_n^2}, \quad \text{(2nd-order)},
\end{equation}
or  
\begin{equation}
  H(s) = \frac{1}{\tau s + 1}, \quad \text{(1st-order)},
\end{equation}
where $\omega_n$ is the natural frequency, $\zeta$ the damping ratio, and $\tau$ the time
constant. Servo-driven aerodynamic surfaces use the second-order form ($\omega_n = 10$~rad/s,
$\zeta = 1/\sqrt{2}$), while propulsion uses the first-order form, with the time constant set by
the powerplant class. The commanded input is clipped to $[-1,1]$, affinely scaled to the physical
range, and the integrated state is saturated at the configured deflection and throttle limits
(e.g. $\pm 20^\circ$ elevator, $\pm 15^\circ$ ailerons and rudder, throttle $\in [0,1]$); both
explicit Euler and RK45 integration of the actuator states are supported, and
the same equations are shared by the CPU reference and GPU-batched backends. Thrust follows
propeller momentum theory, $T = \tfrac{1}{2}\rho S_p C_p V_d (V_d - V_a)$ with slipstream velocity
$V_d = V_a + \delta_t (k_m - V_a)$, whose constants are identified per airframe: from
motor/propeller data for the electric platforms, and from rated shaft power and propeller geometry
(static momentum-disk thrust plus power-available cruise thrust) for the piston-engine
general-aviation airframes. Powerplant class therefore enters the model both through this
identification and through the throttle lag: electric motors respond fast ($\tau = 0.2$~s),
whereas piston engines are markedly slower ($\tau = 3$~s), which is the dominant throttle-loop
latency for those aircraft. This modeling approach is commonly adopted in aircraft simulation
studies \cite{nelson1998flight}. The framework is implemented in a modular, factory-registered
manner, so alternative propulsion systems (e.g. turbofans) can be added with minimal structural
changes. Asymmetric thrust configurations are supported: per-motor throttle states, mounting
positions, and rotation directions produce differential yaw and roll moments through
$\mathbf{r}\times\mathbf{F}$ and the propeller torque reaction, and can be exploited for control.



\textit{Environmental Effects}: Constant wind, Dryden turbulence \cite{dryden1930effect}, and
International Standard Atmosphere density gradients \cite{USStandardAtmosphere1976} (ISO~2533 troposphere, equivalent to the 1976 U.S. Standard Atmosphere below 11~km) are injected into the dynamics, enabling robustness testing under realistic conditions. Constant wind is specified in the inertial frame and rotated into the body frame, where it is summed with the turbulent gusts; the total gust vector perturbs the airspeed used by the aerodynamic model, $\mathbf{V}_a = \mathbf{v} - \mathbf{w}_g$, so that angle of attack, sideslip and dynamic pressure all respond to the disturbance. The Dryden model follows MIL-F-8785C and is realized as band-limited white noise passed through the shaping filters of Table~\ref{tab:dryden}, integrated at the simulation step alongside the rigid-body
states. Scale lengths and intensities are re-evaluated every step at the instantaneous altitude and
airspeed. For low-altitude flight ($h < 1000$~ft), which is the regime of interest here, 
\begin{equation}
 L_u = L_v = \frac{h}{(0.177 + 0.000823h)^{1.2}}, \quad L_w = h,
\end{equation}
and
\begin{equation} 
\sigma_u = \sigma_v = \frac{\sigma_w}{(0.177 + 0.000823h)^{0.4}}, \quad \sigma_w = 0.1\,W_{20},
\end{equation}
while above 1000~ft the isotropic form $L_u = L_v = L_w = h$ is used. Here $h$ is the altitude in
feet and $W_{20}$ the wind speed at 20~ft, which sets the turbulence severity: the presets
\emph{very light}, \emph{light}, \emph{moderate} and \emph{severe} correspond to $W_{20} = 2$, 15,
30 and 45~kt. Wind and turbulence are disabled by default, so training runs on nominal air unless
domain randomization is explicitly requested, and are enabled at evaluation time to quantify
disturbance rejection (Section~V). The CPU reference and Torch/Warp backends implement the identical 
model.

\begin{table}[h!]
\centering
\caption{Dryden turbulence velocity spectral filters.}
\label{tab:dryden}
\resizebox{\columnwidth}{!}{%
\begin{tabular}{@{}cccc@{}}
\toprule
 & \textbf{Longitudinal} & \textbf{Lateral} & \textbf{Vertical} \\
\midrule
Filter &
$\displaystyle G_{u}(s)=\dfrac{\sigma_{u}K_{u}}{(1+T_{u}s)^{2}}$ &
$\displaystyle G_{v}(s)=\dfrac{\sigma_{v}K_{v}(1+\sqrt{3}T_{v}s)}{(1+T_{v}s)^{2}}$ &
$\displaystyle G_{w}(s)=\dfrac{\sigma_{w}K_{w}(1+\sqrt{3}T_{w}s)}{(1+T_{w}s)^{2}}$ \\[10pt]
Constants &
$K_{u} = \sqrt{\dfrac{2L_{u}}{\pi U_{0}}},\ T_{u} = \dfrac{L_{u}}{U_{0}}$ &
$K_{v} = \sqrt{\dfrac{L_{v}}{\pi U_{0}}},\ T_{v} = \dfrac{L_{v}}{U_{0}}$ &
$K_{w} = \sqrt{\dfrac{L_{w}}{\pi U_{0}}},\ T_{w} = \dfrac{L_{w}}{U_{0}}$ \\
\bottomrule
\end{tabular}%
}
\end{table}

\begin{figure}[ht!]
    \centering
    \begin{subfigure}[t]{0.48\columnwidth}
        \centering
        \includegraphics[width=\linewidth]{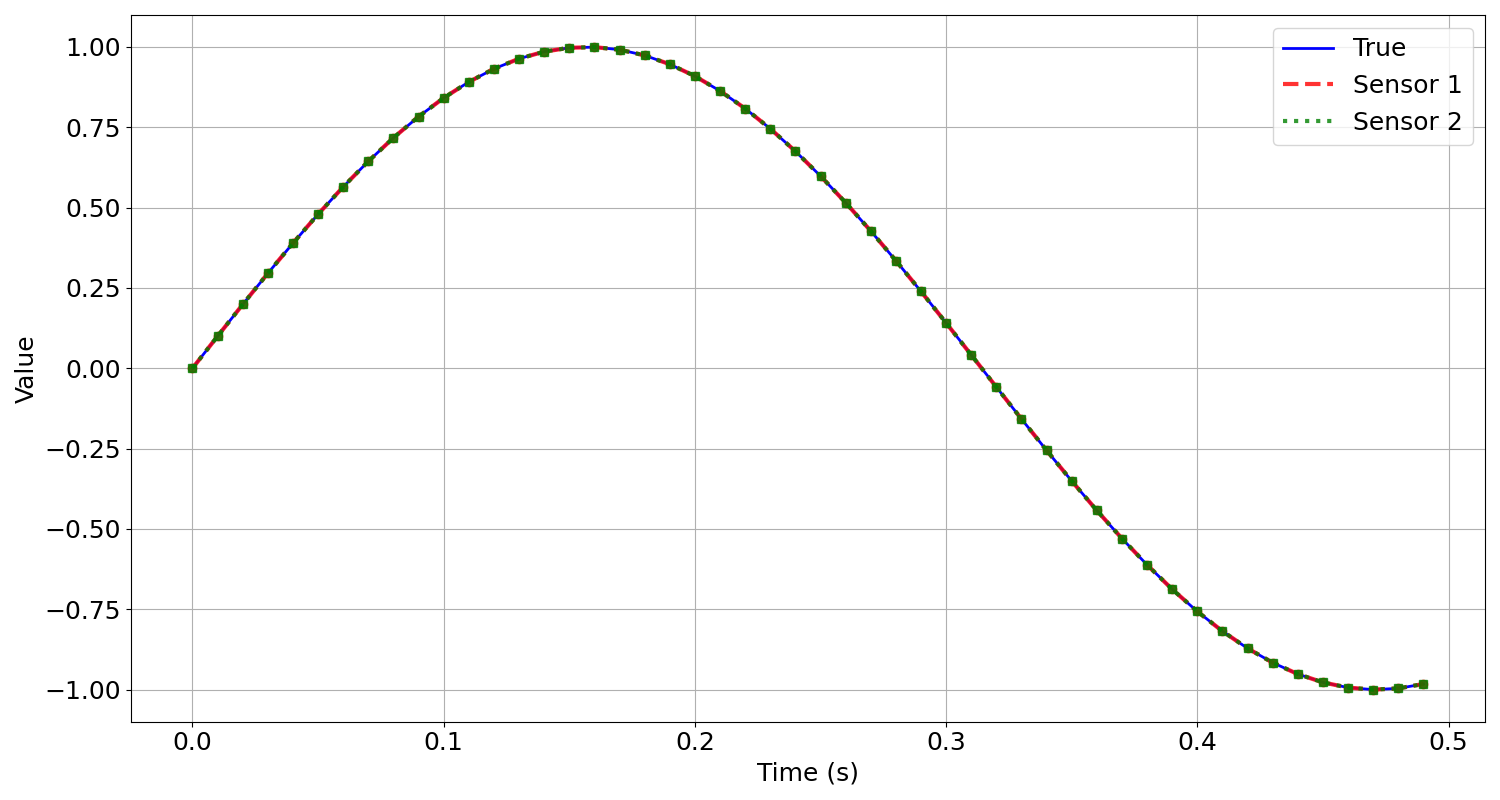}
        \caption{Perfect}
    \end{subfigure}
    \hfill
    \begin{subfigure}[t]{0.48\columnwidth}
        \centering
        \includegraphics[width=\linewidth]{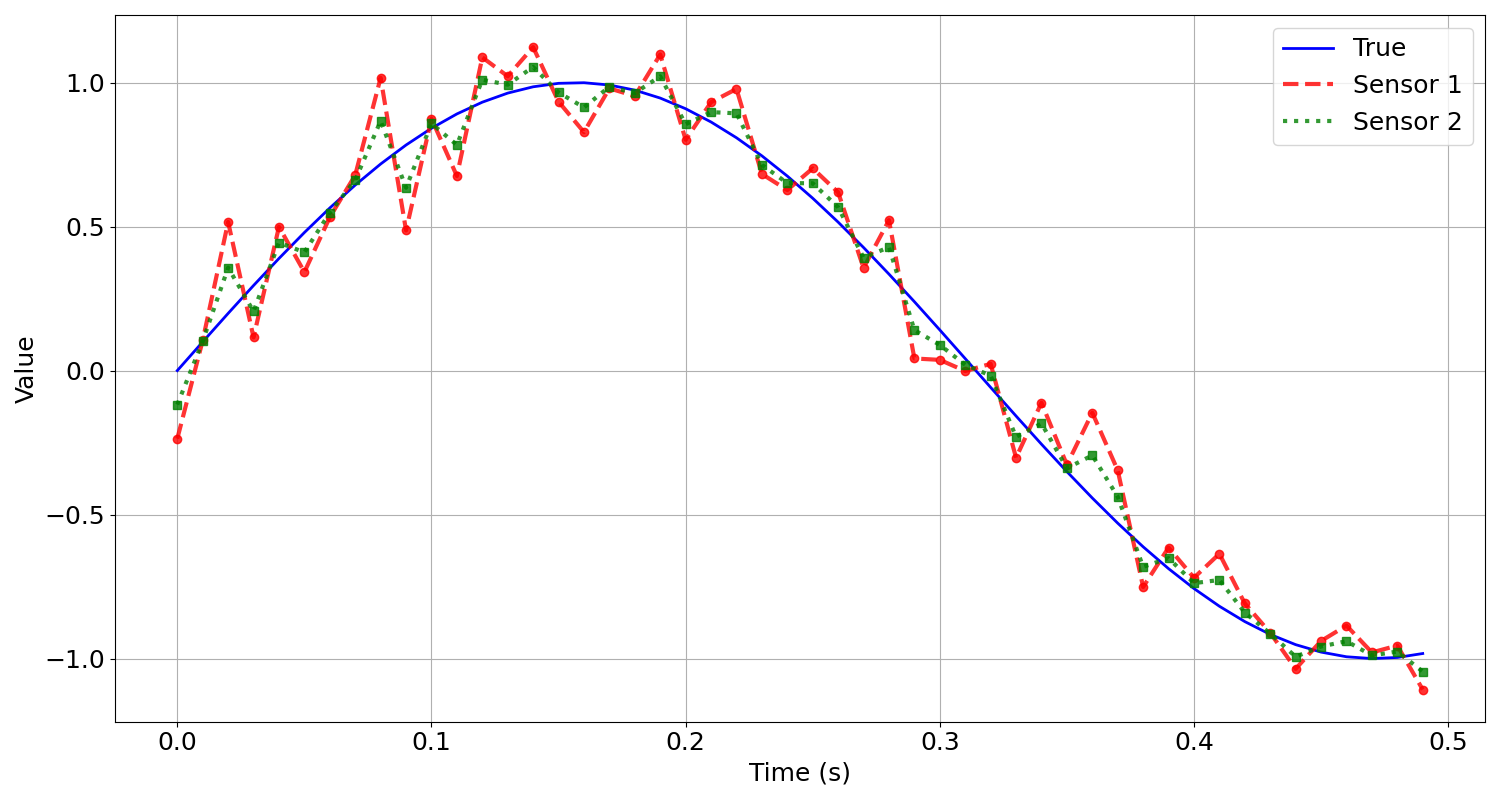}
        \caption{Noise}
    \end{subfigure}
    \hfill
    \begin{subfigure}[t]{0.48\columnwidth}
        \centering
        \includegraphics[width=\linewidth]{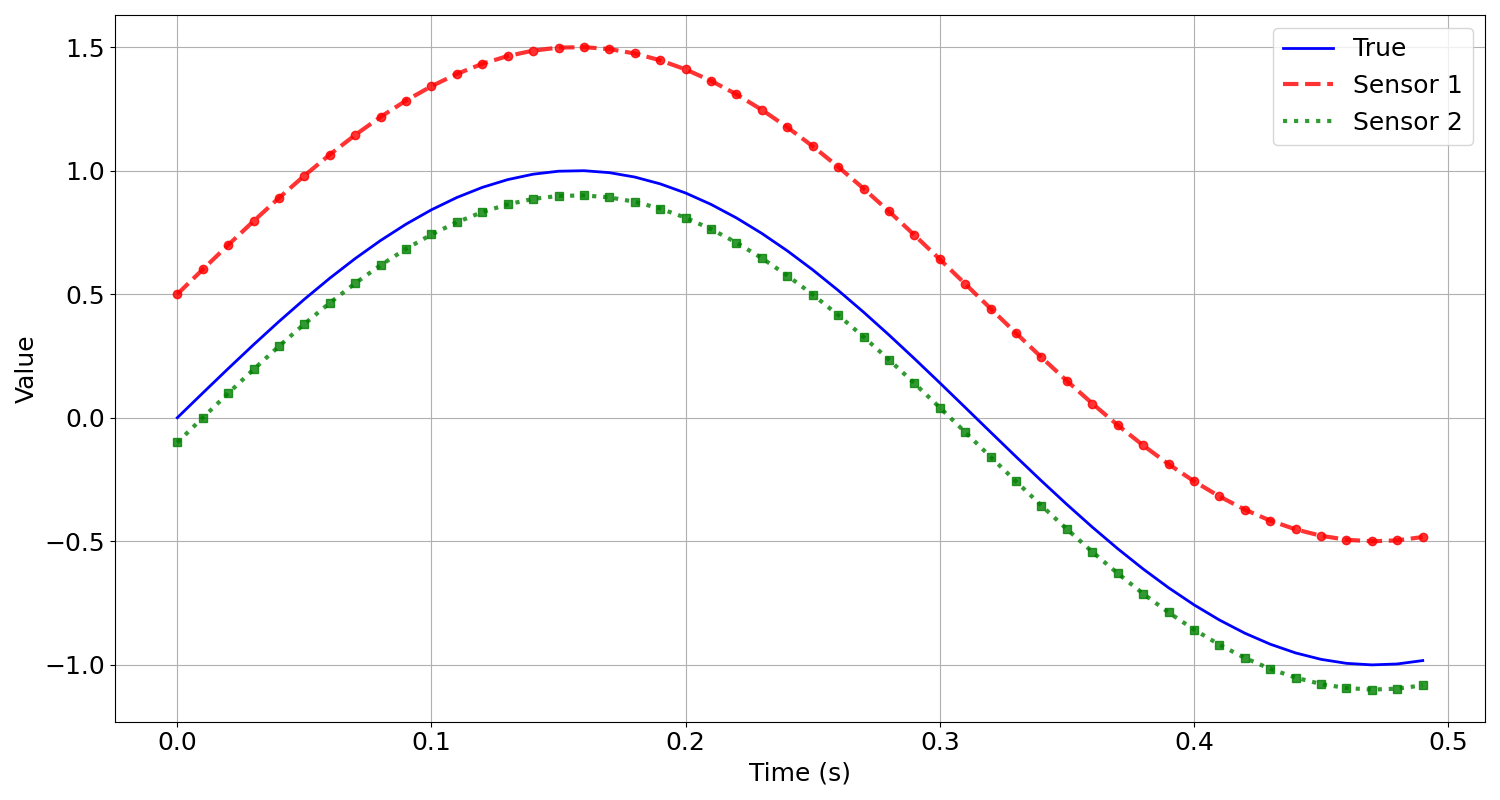}
        \caption{Bias}
    \end{subfigure}
    \hfill
    \begin{subfigure}[t]{0.48\columnwidth}
        \centering
        \includegraphics[width=\linewidth]{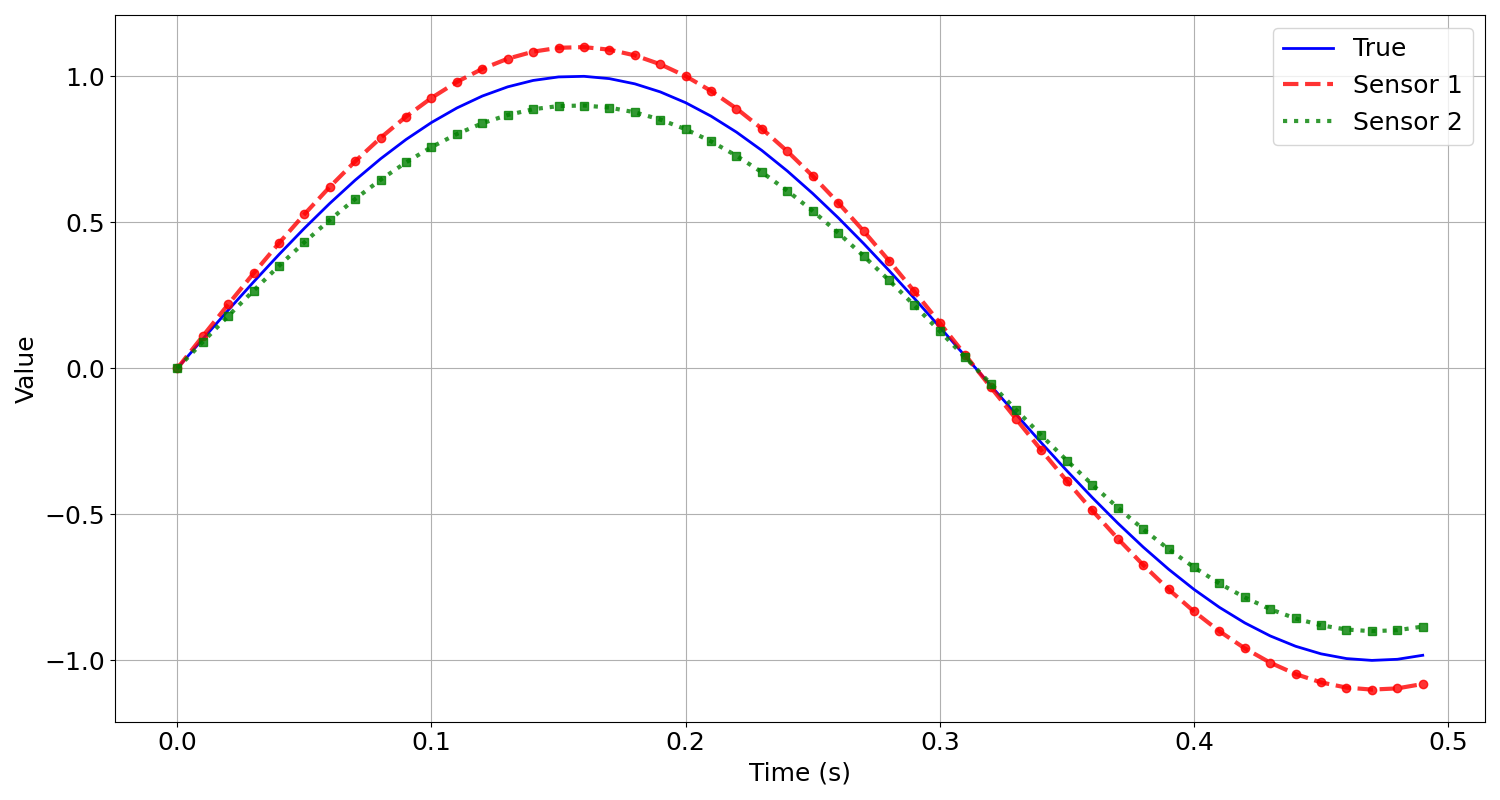}
        \caption{Scaling}
    \end{subfigure}
    \caption{Illustration of sensor imperfections supported by the simulator, including noise, bias and scaling.
    }
    \label{fig:sensor_effects}
\end{figure}

\textit{Sensors}: An onboard sensor model exposes the quantities a flight-control stack
actually consumes: GPS position and velocity, vehicle attitude, and body angular rates from the
gyroscope. Each channel is configured independently and can be degraded by additive Gaussian
noise, constant bias, scale-factor error, clipping to sensor limits, quantization at a finite
resolution, reduced sampling rate (zero-order hold between samples), and transport delay
(implemented as a timestamped circular buffer, so a measurement is released only after the
configured latency has elapsed). Fig.~\ref{fig:sensor_effects} illustrates these effects on sensor outputs
compared to the ground-truth signal. The attitude channel is treated specially: only noise and
limits are applied and the quaternion is renormalized afterwards, so that the measurement remains a
valid rotation. Sensor routing is opt-in: policies observe the ground-truth state by default, and
the sensor chain is enabled to quantify degradation under noisy, rate-limited and delayed sensing
(Section~V), or to match the estimator output available on the target vehicle in preparation for
sim-to-real transfer. Each disturbance can be toggled or combined, and the CPU reference and Warp
backends implement the identical sensor models.

  \textbf{Flight Tasks}: The agent interacts with the environment through a set of modular task
  definitions, such as fixed-altitude keeping, dynamic climbing/descending, 2D path following, full 3D
  trajectory tracking, and attitude-command tracking (bank angle, climb rate, and airspeed references).
  These are defined as reward functions and success conditions on top of the raw physics simulation. Each physics component can be independently toggled or simplified, enabling ablation studies under
  controlled settings. 

\textbf{Aircraft 3D model}:
Our framework supports rapid prototyping of different airframes via JSON-based configuration files. Each aircraft model (e.g., the Airship V7 and A0S WIG prototypes, Volantex Ranger, Cirrus SR22, or Navion) is described by its geometry, mass, inertia, control surface layout, propulsion system parameters, sensor configuration and environmental settings. Given the aircraft OpenVSP 3D model, using its python API, the aerodynamic coefficients can be computed as a look-up table and then fitted to a $N$th order polynomial, as a function of control surface deflection, angle of attack and sideslip angle. These models support both simulation and visualization, and their modular design enables easy switching between vehicles and configurations to evaluate control policies.

\subsection{GPU Backend Performance}
\begin{figure}[ht!]
  \centering
  \includegraphics[width=\linewidth]{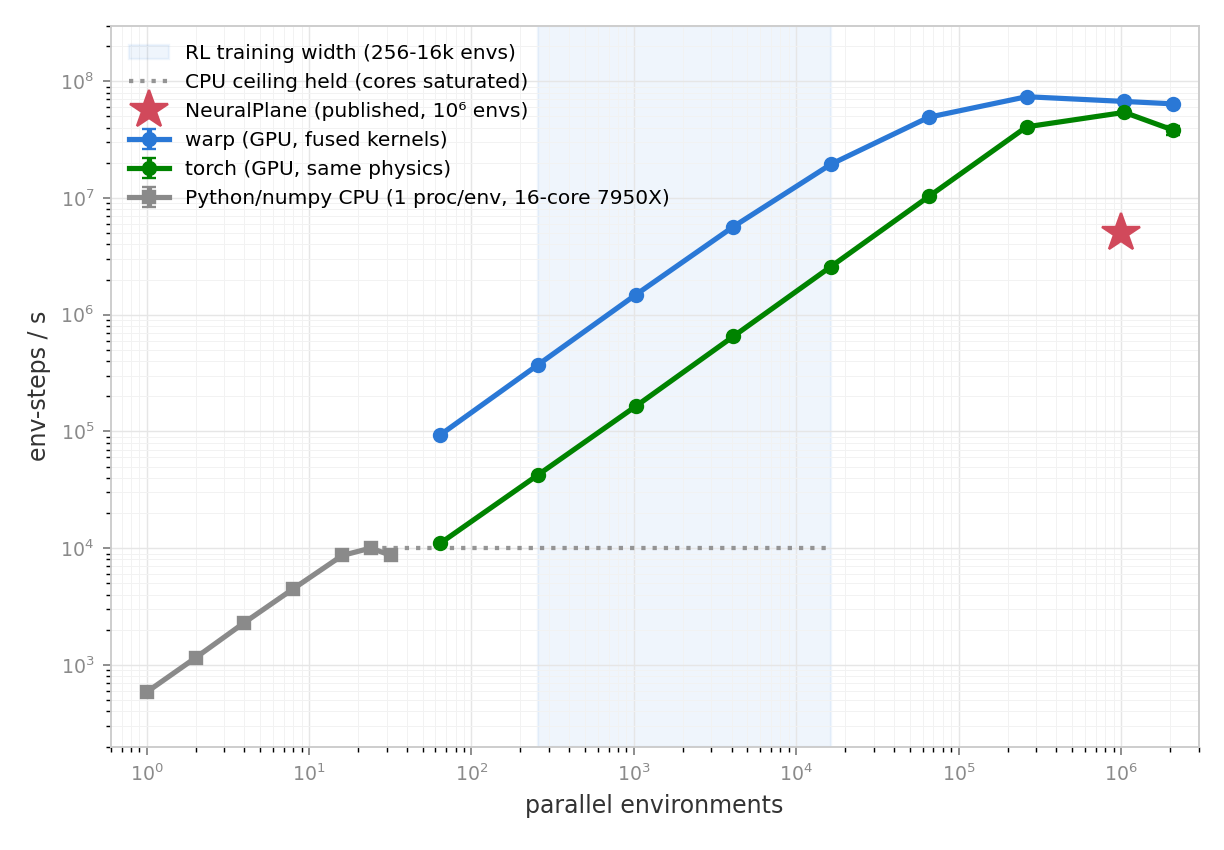}
  \caption{\small Throughput vs.\ parallel environments for the Warp, PyTorch, and CPU backends (log-log).}
  \label{fig:throughput_scaling_c}
\end{figure}

We benchmark the three backends on identical Airship~V7 altitude-task dynamics (with same
aerodynamics, actuator models, ground effect and time step) on an NVIDIA RTX~4090 and an AMD
Ryzen~9~7950X, averaging 10 CUDA-fenced trials after warm-up exclusion (relative standard deviation
$\leq 0.3\%$). Table~\ref{tab:throughput} and Fig.~\ref{fig:throughput_scaling_c} summarize the results. To
keep the CPU baseline honest it is parallelized the way CPU RL stacks are, one process per
environment: it scales near-ideally to the 16 physical cores and saturates at $10$k env-steps/s,
beyond which the aggregate is flat by construction.

The Warp backend scales near-linearly across the widths RL training actually uses
($4$k--$16$k environments), reaching $19.5$M env-steps/s at $16$k (three orders of magnitude
above the saturated CPU machine, and $7.5\times$ the parity-validated PyTorch backend) and peaks
near $2 \times 10^{5}$ environments before becoming bandwidth-bound. At $10^{6}$ environments it
sustains $67.1$M env-steps/s, $13.4\times$ the figure reported for NeuralPlane~\cite{xue2024neuralplane} at the same width. Simulation therefore ceases to be the bottleneck for learning: it accounts for under $10\%$ of training wall-clock, which is instead set by gradient updates (Section~V).
Our framework allows a PPO altitude policy to perform $150$ million environment steps in under five minutes of wall-clock time, while also supporting efficient sampling for methods such as Model Predictive Path Integral (MPPI) control and large-scale parallel evaluation.

\begin{table}[ht!]
\centering
\caption{Simulation throughput by backend and parallel-environment count (mean over 10 trials).}
\label{tab:throughput}
\small
\setlength{\tabcolsep}{5pt}
\resizebox{\columnwidth}{!}{%
\begin{tabular}{lrrr}
\toprule
\textbf{Backend} & \textbf{Envs} & \textbf{Env-steps/s} & \textbf{Speedup vs.\ CPU} \\
\midrule
Python/SciPy (CPU) & 1        & 589            & $1\times$ \\
PyTorch (GPU)      & 16{,}384 & 2{,}586{,}610  & $4{,}390\times$ \\
Warp (GPU)         & 16{,}384 & \textbf{19{,}461{,}717} & $\mathbf{33{,}029\times}$ \\
Warp (GPU)         & 262{,}144 & \textbf{73{,}640{,}000} & peak \\
Warp (GPU)         & 1{,}048{,}576 & 67{,}070{,}000 & saturated \\
\bottomrule  
\end{tabular}
}
\end{table}

\subsection{Cross-Simulator Validation}

We validate our simulator against two independent environments: X-Plane and JSBSim.

\textit{X-Plane}: X-Plane's flight dynamics are closed, so a term-by-term model comparison is
not possible; instead we test controller transfer. Through the Python XPlaneConnect
API~\cite{xplaneconnect}, a controller tuned in our simulator is closed around the X-Plane aircraft
reading its state and writing elevator, aileron, rudder and throttle commands at run time, from the
same configuration and initial conditions. We consider retaining stable tracking under an independent,
industry-standard physics engine to be qualitative evidence of robustness rather than a numerical
validation. The same interface renders high-fidelity footage of trained agents (available in the
open-source repository), allowing for high quality video-feedback.

\textbf{JSBSim}: For quantitative evidence of physics fidelity, we cross-validate against
JSBSim~\cite{berndt2004jsbsim} under identical zero-input free-flight conditions at $h_0 \in \{1, 10, 100\}$\,m (Table~\ref{tab:jsbsim_corr}). With ground effect disabled, trajectories agree near-perfectly (Pearson $r \geq 0.999$ on altitude and pitch rate throughout), confirming baseline dynamic consistency. Enabling ground effect at $h_0 = 1$\,m produces a clear, systematic divergence (physically expected, since JSBSim does not model ground effect) directly isolating our GE correction as a real aerodynamic phenomenon rather than a numerical artefact; the lower airspeed correlation in this condition ($r = 0.948$) reflects precisely the altered drag dynamics JSBSim does not replicate. 

\begin{table}[ht!]
\centering
\caption{\small Pearson correlations between our simulator and JSBSim under identical
zero-input free-flight conditions. GE: ground effect.}
\label{tab:jsbsim_corr}
\small
\setlength{\tabcolsep}{4pt}
\begin{tabular}{lccc}
\toprule
\textbf{Condition} & \textbf{Altitude} & \textbf{Pitch rate} & \textbf{Airspeed} \\
\midrule
$h_0 = 1$\,m, GE on  & 0.9999 & 0.9989 & 0.9480 \\
$h_0 = 1$\,m, GE off & 0.9999 & 0.9999 & 0.9999 \\
$h_0 = 10$\,m        & 1.0000 & 0.9885 & 0.9997 \\
$h_0 = 100$\,m       & 0.9996 & 0.9975 & 0.9973 \\
\bottomrule
\end{tabular}       
\end{table}

\subsection{Visualization Tools}

A set of Python-based visualization tools, built primarily on matplotlib~\cite{matplotlib}, supports analysis of simulation outputs: time-domain plots of all model states, aerodynamic forces, moments and coefficients, actuator commands, and ground-effect indicators, alongside animated 3D flight-path renderings with an attitude-resolved aircraft glyph for qualitative assessment of maneuvers (videos in the open-source repository). These tools generate the result figures throughout Section~\ref{sec:experiments} (e.g., maneuver-tracking and learning-curve figures), while the X-Plane interface provides photorealistic renderings such as Fig.~\ref{fig:hero}.


\section{Experiments \& Results}\label{sec:experiments}

Our experiments quantify how modelling choices and task conditions affect controller performance,
and compare representative paradigms to provide a usable flight control benchmark. We evaluate
five controllers, two classical (LQR, MPPI) and three learned (PPO, SAC, TD3), on two
tasks: holding a commanded altitude after acquiring it from a known offset, and tracking a stream
of attitude commands through four different maneuvers. Both tasks run on the same two reference
airframes: the Airship~V7, the ground-effect prototype the framework was built around, and the
Volantex Ranger, a widely used remote-controlled electric RC vehicle.

We then expand the experiments to analyse the influence of Ground-Effects on flying at different altitudes (section~\ref{sec:ge-energy} and tests under environment variations, quantifying how controlled flights degrade under sensor noise and turbulence (section~\ref{sec:robustness}). Every cell of every comparison runs on identical conditions, i.e. same airframe configuration, same actuator and sensor models, same reference stream.

\subsection{Task Definitions} \label{subsec:tasks}

Each task is a Markov decision process over the 6DoF plant of Section~\ref{sec:preliminaries},
integrated at $\Delta t = 10$~ms, with observations normalised by fixed physical scales and
clipped, and actions in $[-1,1]$ mapped affinely onto the physical actuator ranges and passed
through the servo and motor dynamics of Eqs.~(6)--(7). Two task families are used.

\emph{Altitude keeping.} The agent regulates height above the surface to a commanded altitude
$h^\star$ while holding trim airspeed. The observation is
\begin{equation}
o = \Big[\tfrac{e_h}{25},\, \tfrac{q}{2},\, \tfrac{e_{V_a}}{\sigma_V},\, \tfrac{\theta}{\pi/4},\,
  \tfrac{v_z}{\sigma_v},\, \tfrac{\alpha}{20^\circ},\, \bar\delta_e,\,
  \tfrac{\dot\delta_e}{100},\, \bar\delta_t,\, \tfrac{\int\! e_h}{10},\,
  \tfrac{\dot h_{\mathrm{ref}}}{5}\Big] \in \mathbb{R}^{11}, 
\end{equation}
clipped elementwise to $[-3,3]$, where $e_h = h - h^\star$, $e_{V_a} = V_a - V_a^\star$, $q$ is
pitch rate, $\bar\delta_e,\bar\delta_t$ are the normalised elevator and throttle states, and
$\dot h_{\mathrm{ref}}$ is the commanded climb rate of the reference. The action is
$a = [\delta_e, \delta_t] \in [-1,1]^2$. The reward combines a Gaussian altitude term with
airspeed, smoothness and safety shaping, 
\begin{equation}
\begin{aligned}
r = {}& w_h \exp\!\big(-e_h^2/2\sigma_h^2\big)
- w_V \tfrac{|e_{V_a}|}{V_a^\star}
- w_q \min(q^2, \bar q)\\
&- w_s \|\Delta a\|\,\rho(e_h)
+ w_h\,\kappa_h \kappa_V
+ b_{\mathrm{surv}}
- w_o\,\phi(e_h),
\end{aligned}
\end{equation}

where $\rho$ gates the action-rate penalty near the reference, $\kappa_h,\kappa_V$ are linear
tapers on altitude and airspeed error that pay a bonus only inside both bands,
$b_{\mathrm{surv}}$ is a per-step survival bonus and $\phi$ penalises overshoot above the target.
An episode terminates on ground contact, on stall ($\alpha$ beyond
$1.5\,\alpha_{\mathrm{stall}}$ or $V_a$ below the stall threshold) or at the horizon.

\emph{Attitude-based tracking.} The agent is a low-level executor: it receives a stream of
bank, climb-rate and airspeed commands $(\phi^\star, \dot h^\star, V_a^\star)(t)$ and owns all
four actuators directly, with no autopilot in between. The observation is
\begin{equation} 
o = \big[e_\phi,\, e_{\dot h},\, e_{V_a},\, \phi,\, \theta,\, p,\, q,\, r,\, \alpha,\, \beta,\,
  v_z,\, a_{t-1}\big] \in \mathbb{R}^{15},
\end{equation}
scaled and clipped to $[-5,5]$, and the action is
$a = [\delta_e, \delta_a, \delta_r, \delta_t] \in [-1,1]^4$. The reward is a sum of Gaussian
tracking terms on the three commanded channels, an airspeed-margin penalty and an action-rate
penalty,
\begin{equation}
r =
\sum_{i\in\{\phi,\dot h,V_a\}}
w_i e^{-e_i^2/(2\sigma_i^2)}
-w_E\left(\frac{V_{\mathrm{safe}}-V_a}{V_{\mathrm{safe}}}\right)^2
-w_s\|\Delta a\|^2 .
\end{equation}
Commanding climb rate rather than pitch attitude is deliberate: pitch attitude does not determine
altitude rate at varying airspeed, so a pitch-tracking executor cannot be composed into an
altitude-holding outer loop.


\subsection{Metrics}

  Every rollout is scored with the same set (Table~\ref{tab:metrics}). \emph{Survival} takes
  precedence over every error metric: error on a controller that has left the flight envelope is not
  meaningful, so errors average only the runs that stayed inside, and a cell where none did is left
  empty. \emph{RMSE} and \emph{time-in-band} measure steady-state accuracy, as a magnitude and
  against the task tolerance. \emph{Acquisition}, \emph{settling time} and \emph{overshoot} measure
  the transient, and exist only because every episode starts a known distance off the reference
  (Section~\ref{sec:altitude-protocol}).

  \begin{table}[t]
  \caption{\small Evaluation metrics computed from trajectory and control logs. $e_t$ is the tracking error
  at step $t$, and $\delta$ the capture band, $1$~m on the altitude task.}
  \label{tab:metrics}
  \centering
  \begin{tabular}{ll}
  \toprule
  Metric & Formula / description \\
  \midrule 
  Survival      & fraction of episodes reaching the horizon without \\
                & crash or stall termination \\
  Acquisition   & fraction additionally reaching and holding the \\
                & target to within $\delta$ \\
  RMSE          & $\sqrt{\frac{1}{|W|}\sum_{t \in W} e_t^2}$ over the settled window $W$ \\
  Time-in-band  & fraction of scored time with $|e_t|$ inside the \\
                & task tolerance \\
  Settling time & $\min t$ s.t. $|e_\tau| \le \delta$ for all $\tau \ge t$ \\
  Overshoot     & $\max_t\, \mathrm{sgn}(h^\star - h_0)\,e_t$, clipped at 0 --- the \\
                & excursion \emph{past} the target, not $\max_t |e_t|$ \\
  \bottomrule  
  \end{tabular}
  \end{table}


\begin{figure*}[t]
\centering
\includegraphics[width=0.49\textwidth]{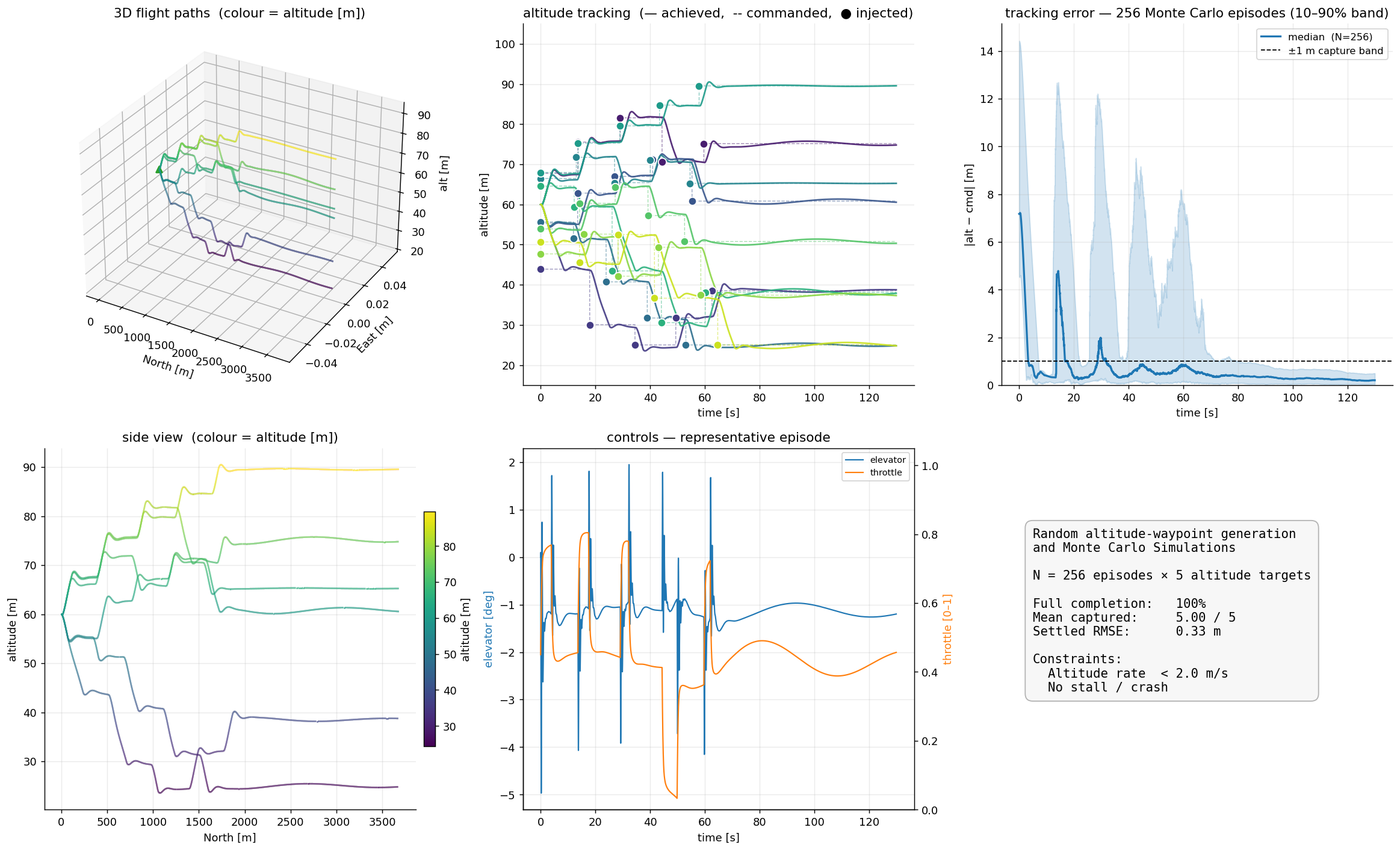}
\hfill
\includegraphics[width=0.49\textwidth]{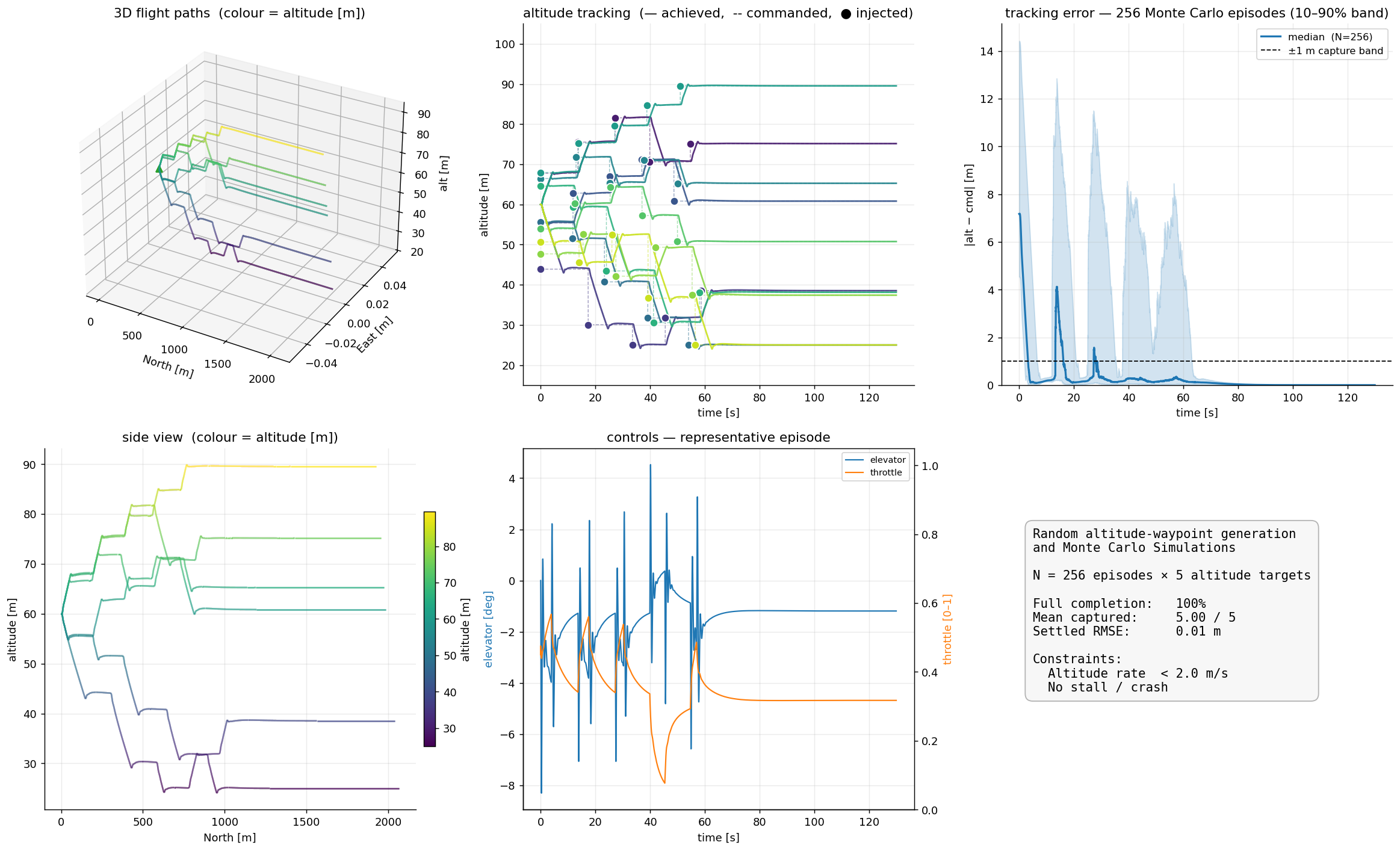}
\caption{PPO holding a sequence of altitude targets, Airship V7 (left) and Volantex Ranger
(right). Each panel set shows the flight paths, commanded versus achieved altitude for ten
episodes, the tracking error over 256 Monte Carlo episodes with its 10--90\% band, a side view,
and the elevator and throttle traces of one episode. A new target is released only once the
current one is held, so each error spike is one acquisition.}
\label{fig:altitude-ppo}
\end{figure*}

\subsection{Altitude Acquisition and Hold}
\label{sec:altitude-protocol}

The first task is to reach a commanded altitude and hold it. Each episode is initialized at a
distance $\Delta\sim\pm\,\mathcal{U}(12,18)$~m from the target, for commanded altitudes of 20, 40,
60, 80 and 100~m, and all five controllers follow the identical reference: a sub-target that ramps
to the commanded altitude at the rate limit of the airframe and then holds. The offset initial
condition creates the acquisition transient, and therefore makes the metrics of
Table~\ref{tab:metrics} meaningful.
Table~\ref{tab:altitude-protocol} summarizes the $n$ episodes of each cell. Survival and acquisition (with $\delta$ equal to 1~m) are reported side by side
because they are not equivalent: MPPI on the Airship~V7 survives 0.93 while acquiring 0.01, and
survival alone would credit it with almost the whole task.
PPO is the only method that survives and acquires every episode on both airframes. SAC and TD3 lose
between 7\% and 17\% of theirs. SAC on the Airship~V7 shows the largest gap between the two rates,
surviving 0.83 while acquiring 0.50: half of its runs stay inside the flight envelope while drifting
away from the commanded altitude. Accuracy is a separate axis. Where it arrives, the LQR is by far
the most precise method, with a settled error of $2.8\times10^{-7}$~m on the Airship~V7 and
$1.4\times10^{-4}$~m on the Volantex Ranger, five and one order of magnitude below the best learned
policy on the same airframe. On a constant reference the task reduces to the linear regulation
problem its gains were computed for, and the residual is numerical rather than a control error. The
two axes do not order the methods in the same way: that same LQR acquires only half of the episodes
on the Volantex Ranger, where the learned policies acquire nearly all.
MPPI acquires the target on neither airframe. On the Airship~V7 it survives 93\% of the episodes and
acquires 1\%, with a settled error of 30.7~m and an overshoot of 41.5~m against a commanded change
of at most 18~m. On the Volantex Ranger it survives 18\% and acquires none. The overshoot says that
the planner flies through the target instead of stopping at it. Its cost is a weighted sum of
absolute position and airspeed errors over a one-second preview, with no term on the vertical speed,
so the climb or sink rate it has built up stays invisible until the aircraft is already past the
target. The failure lies in the cost and in the preview length rather than in the sampling scheme:
the same planner holds three of the four V7 attitude maneuvers of Section~\ref{sec:maneuvers} below
$0.7^\circ$, where the commanded quantity is itself a rate. MPPI also picks its plan by scoring a
batch of random candidate control sequences, so each sweep is one draw. On a single Volantex Ranger
cell the settled error ranges over 2.6--15.8~m across repetitions of the identical setup, and its
row is a mean over three independent sweeps.
The learned rows are seed averages, and the spread is as informative as the mean. SAC on the
Airship~V7 acquires $0.50\pm0.41$ with a settled error of $3.52\pm3.55$~m, a standard deviation as
large as the mean itself, against $1.00\pm0.00$ and $0.288\pm0.022$~m for PPO. Scoring a single
checkpoint would have placed SAC either above or below PPO depending on the run.
Settling time lies between 6.9 and 10.0~s for every method that arrives. It is set by the rate limit
of the reference rather than by the controller, so it does not discriminate here, and it is left
empty for MPPI, whose acquisition rate is too low for it to mean anything. The results should also
be read against the per-airframe design each method requires: gains derived for the acquisition
condition and a reference airspeed set to trim for the classical controllers, a single training run
from the same configuration file for the learned policies.
Figure~\ref{fig:altitude-ppo} shows PPO flying a sequence of altitude targets, each released only
once the previous one is held. The tracking error collapses inside the $\pm1$~m capture band after
every change of target and remains there, with elevator activity confined to the transients and the
throttle carrying the steady-state trim between them. 

\begin{table}[t]
\caption{\small Altitude acquisition and hold. Surv. is survival rate, Acq. is successful acquisition, RMSE is altitude error, $t_s$ is settling time, and Over. is overshoot.}
\label{tab:altitude-protocol}
\centering
\small
\setlength{\tabcolsep}{3.pt}
\begin{tabular}{@{}llrrrrrr@{}}
\toprule
& Alg. & $n$ & Surv. & Acq. & RMSE (m) & $t_s$ (s) & Over. (m) \\
\midrule
\multirow{5}{*}{\rotatebox{90}{Airship V7}}
& PPO  & 200 & 1.0$\pm$0.0 & 1.0$\pm$0.0 & 0.3$\pm$0.0 & 7.8$\pm$0.4 & 0.8$\pm$0.0 \\
& SAC  & 200 & 0.8$\pm$0.2 & 0.5$\pm$0.4 & 3.5$\pm$3.6 & 6.9$\pm$0.7 & 0.2$\pm$0.1 \\
& TD3  & 200 & 0.9$\pm$0.2 & 0.9$\pm$0.2 & 0.1$\pm$0.1 & 8.3$\pm$1.1 & 0.6$\pm$0.4 \\
& LQR  & 50  & 1.0 & 1.0 & $<0.1$ & 10.0 & 1.9 \\
& MPPI & 50  & 0.9$\pm$0.0 & 0.0$\pm$0.0 & 30.7$\pm$0.3 & --- & 41.5$\pm$0.4 \\
\midrule
\multirow{5}{*}{\rotatebox{90}{Volantex}}
& PPO  & 200 & 1.0$\pm$0.0 & 1.0$\pm$0.0 & 0.0$\pm$0.0 & 7.3$\pm$0.0 & 0.7$\pm$0.0 \\
& SAC  & 200 & 0.9$\pm$0.2 & 0.9$\pm$0.2 & 0.0$\pm$0.0 & 7.1$\pm$0.1 & 0.4$\pm$0.0 \\
& TD3  & 200 & 0.9$\pm$0.1 & 0.9$\pm$0.1 & 0.0$\pm$0.0 & 7.1$\pm$0.1 & 0.4$\pm$0.0 \\
& LQR  & 50  & 0.5 & 0.5 & $<0.1$ & 9.6 & 1.4 \\
& MPPI & 50  & 0.2$\pm$0.1 & 0.0$\pm$0.0 & 28.1$\pm$6.6 & --- & 28.6$\pm$5.1 \\
\bottomrule
\end{tabular}
\end{table}


  \begin{figure*}[t]
  \centering
  \includegraphics[width=\textwidth]{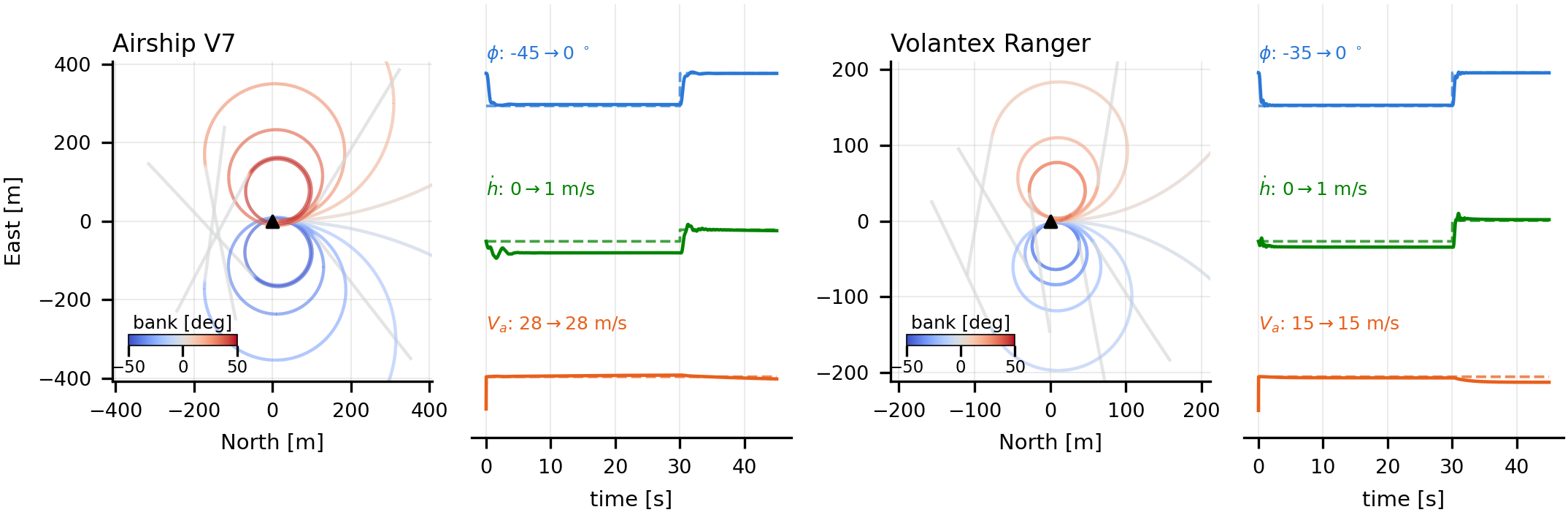}
\caption{\small Attitude-command interface for Airship~V7 (left pair) and Volantex Ranger (right pair). Left: ground tracks across the commanded bank envelopes, coloured by achieved bank. Right: commanded (dashed) and achieved (solid) channels for the steepest sustained turn, normalised by their respective ranges.}

  \label{fig:attitude-exec}
  \end{figure*}

\begin{figure*}[t]
\centering
\includegraphics[width=\textwidth]{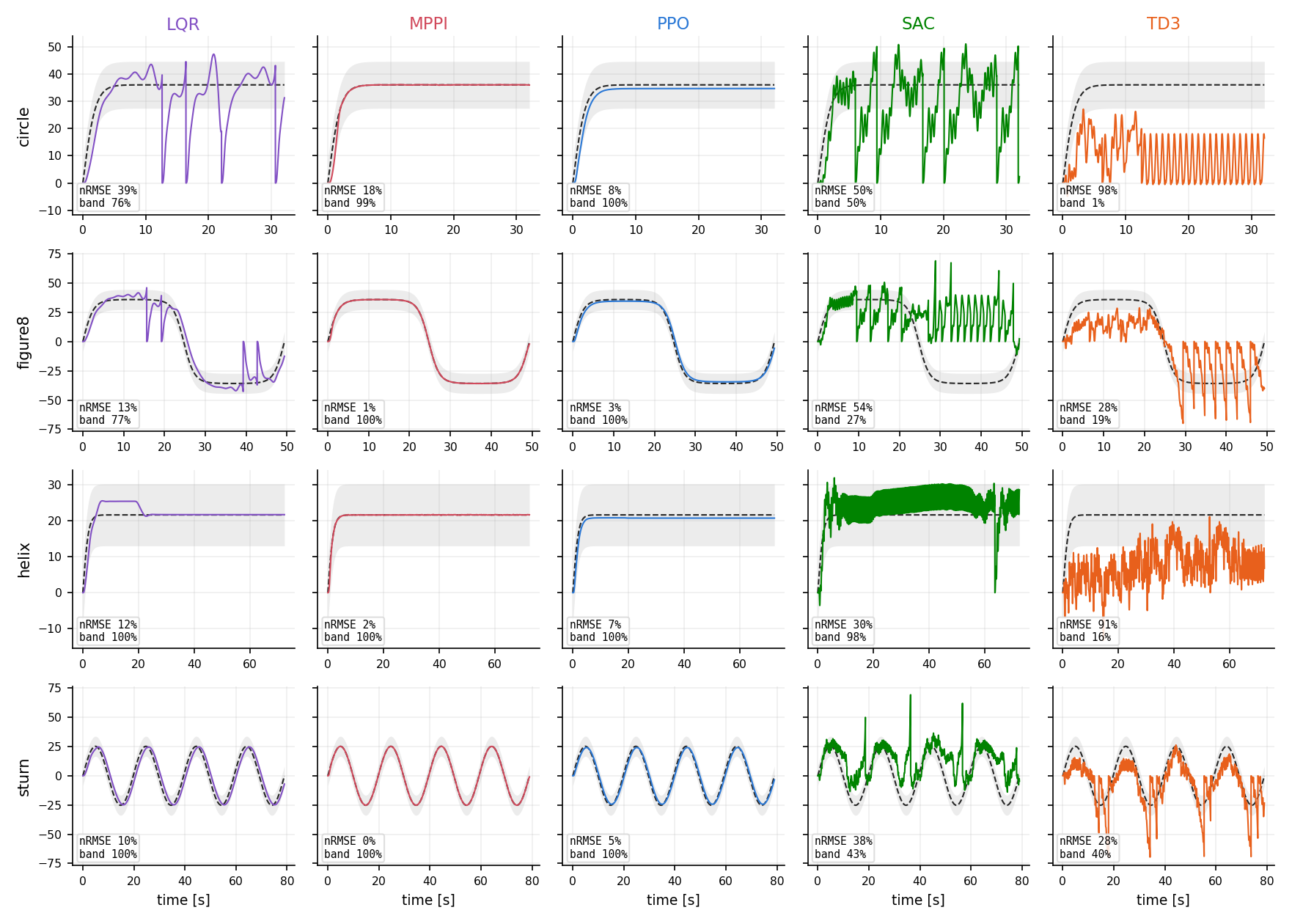}
\caption{\small Bank tracking on Airship~V7. Rows show the four maneuvers; columns show the five executors. Dashed lines show commanded bank, with the $\pm8.6^\circ$ tolerance shaded; panels report RMSE and time-in-band. Faceting reveals LQR departures, SAC tremor, and the TD3 limit cycle, which overlap other traces on shared axes.}
\label{fig:maneuver-facet}
\end{figure*}

  \begin{figure*}[t]
  \centering
  \includegraphics[width=\textwidth]{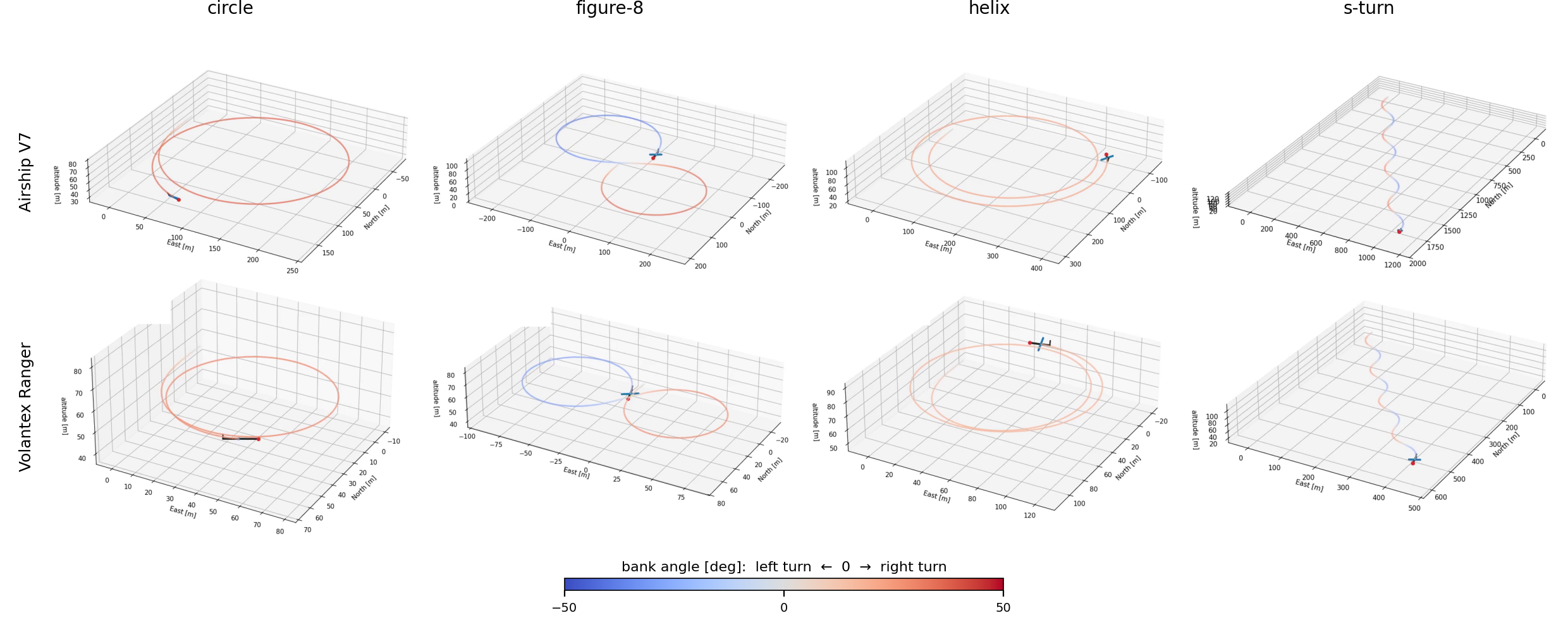}
  \caption{\small Completed maneuvers flown by the attitude executor, rendered from the simulator's 3D
  visualisation. Rows are the two airframes, columns the four maneuvers, and the track is coloured by
  bank angle over $\pm 50^\circ$, blue left and red right. Note the axis scales: a maneuver name
  denotes a comparable difficulty, not a comparable geometry (Table~\ref{tab:maneuver-defs}).}

  \label{fig:maneuver-render}
  \end{figure*}

\subsection{Attitude-based Maneuvers Tracking}
\label{sec:maneuvers}

\begin{table}[t]
\caption{\small Maneuver definitions. Each stream uses the airframe's own envelope, so $\phi_{\max}$, turn radius $R$, and load factor $n=1/\cos\phi_{\max}$ vary between airframes. Commands are identical for all five controllers.}
\label{tab:maneuver-defs}
\centering
\begin{tabular}{llrrr}
\toprule
Airframe & Maneuver & $\phi_{\max}$ ($^\circ$) & $R$ (m) & $n$ \\
\midrule
Airship V7      & circle    & 36.0 & 110 & 1.23 \\
                & figure-8  & 35.8 & 111 & 1.23 \\
                & helix     & 21.6 & 202 & 1.07 \\
                & s-turn    & 25.2 & 170 & 1.10 \\
\midrule
Volantex Ranger & circle    & 28.0 &  43 & 1.13 \\
                & figure-8  & 27.9 &  43 & 1.13 \\
                & helix     & 16.8 &  76 & 1.04 \\
                & s-turn    & 19.6 &  64 & 1.06 \\
\bottomrule
\end{tabular}
\end{table}

\begin{table}[t]
\caption{\small Bank-angle RMSE (deg), mean $\pm$ standard deviation, with
time-in-band (\%) in brackets. Out-of-envelope runs are excluded.}
\label{tab:maneuver}
\centering
\small
\setlength{\tabcolsep}{1.pt}
\begin{tabular}{@{}llrrrr@{}}
\toprule
& Alg. & Circle & Fig.-8 & Helix & S-turn \\
\midrule
\multirow{5}{*}{\rotatebox{90}{Airship V7}}
& PPO  & 1.8$\pm$0.2 (100) & 2.2$\pm$0.1 (100) & 0.8$\pm$0.2 (100) & 2.3$\pm$0.1 (100) \\
& SAC  & 13.0$\pm$1.4 (49) & 20.9$\pm$12.8 (47) & 7.1$\pm$2.4 (83) & 12.7$\pm$4.9 (58) \\
& TD3  & 23.6$\pm$6.4 (18) & 18.0$\pm$5.6 (33) & 11.1$\pm$3.2 (49) & 12.1$\pm$2.1 (49) \\
& LQR  & 10.6 (76) & 9.1 (77) & 2.0 (100) & 5.2 (100) \\
& MPPI & 4.8$\pm$6.9 (88) & 0.7$\pm$0.3 (100) & 0.3$\pm$0.2 (100) & 0.2$\pm$0.1 (100) \\
\midrule
\multirow{5}{*}{\rotatebox{90}{Volantex}} 
& PPO  & 1.0$\pm$0.0 (100) & 1.8$\pm$0.0 (100) & 0.4$\pm$0.0 (100) & 2.0$\pm$0.0 (100) \\
& SAC  & 6.7$\pm$3.4 (74) & 7.2$\pm$1.8 (77) & 3.9$\pm$1.0 (96) & 7.2$\pm$0.9 (75) \\
& TD3  & 5.3$\pm$1.3 (88) & 5.5$\pm$0.8 (87) & 5.3$\pm$2.7 (84) & 5.9$\pm$1.3 (85) \\
& LQR  & 2.0 (100) & 2.4 (100) & 1.6 (100) & 2.5 (100) \\
& MPPI & 0.5$\pm$0.4 (100) & 7.4$\pm$13.7 (89) & 1.6$\pm$1.8 (97) & 19.1$\pm$14.6 (71) \\
\bottomrule
\end{tabular}
\vspace{-.5cm}
\end{table}

The second task family evaluates the attitude executor of Section~\ref{subsec:tasks} on four
maneuvers: a sustained circle, a figure-of-eight, a climbing helix and a weaving s-turn. Each
maneuver is a command stream $(\phi^\star, \dot h^\star, V_a^\star)(t)$ rather than a position-space
path, so the controllers are scored on the attitude and energy states they reach and not on waypoint
tracking. Fig.~\ref{fig:attitude-exec} shows the command interface: bank angles spanning the
airframe envelope, and the three command channels of one turn, held for \SI{30}{\second} before
levelling out and climbing at \SI{1}{\metre\per\second}. Each stream is generated from its own
airframe's envelope. A maneuver name therefore denotes a comparable difficulty, not a comparable
geometry (Table~\ref{tab:maneuver-defs}), and Fig.~\ref{fig:maneuver-render} shows the completed
tracks.
Each learned method is flown once per training seed and reported as the mean over three seeds,
matching the altitude protocol of Section~\ref{sec:altitude-protocol}. The LQR is deterministic and
is flown once. MPPI has no seeds but is flown five times, for the reason given below.
PPO is the only method inside the tolerance band on all eight airframe--maneuver cells
(Table~\ref{tab:maneuver}), at $0.42$--$2.28^\circ$ of bank RMSE, and it is also the most
repeatable: its seed spread never exceeds $0.21^\circ$. One of its three V7 s-turn seeds
nonetheless left the flight envelope, the only PPO departure in the sweep. MPPI is the most
accurate executor where it converges, holding three of the four V7 maneuvers to
$0.2$--$0.7^\circ$. It does not converge everywhere: $4.8^\circ$ on the V7 circle, $7.4^\circ$ on
the Ranger figure-of-eight and $19.1^\circ$ on the Ranger s-turn, where one draw also left the
envelope. The LQR gains are derived about a level trim rather than a banked one. It holds the band
on the Ranger, but only \SI{76}{\percent} and \SI{77}{\percent} of the V7 circle and
figure-of-eight. SAC and TD3 never leave the envelope and track an order of magnitude more loosely
on the V7, to $23.6^\circ$ as a seed mean and $38.9^\circ$ on the worst single seed. This nearly
reverses the altitude ordering of Section~\ref{sec:altitude-protocol}. The planner that cannot
acquire an altitude is the most accurate attitude executor here, where it converges.
A bank-RMSE column hides how each method fails, and the per-executor panels of
Fig.~\ref{fig:maneuver-facet} separate them. On the V7 figure-of-eight SAC under-banks and lags,
with a fitted bank gain of $0.54$ against a $1.2$~s phase lag, and it is the least repeatable of
the learned cells at $20.9\pm12.8^\circ$ across seeds. On the circle it over-banks instead, at a gain of
$1.28$, and loses half of its time-in-band to a $13.0^\circ$-RMS tremor about the command. TD3
settles into a limit cycle on the same circle, at a gain of $0.46$, a $0.6$~s lag and a
time-in-band of \SI{18}{\percent}. The LQR holds the command closely between brief departures,
where the linearisation about level trim loses the banked state. On shared axes the two off-policy
traces cover the other three entirely, hence the facet.
The remaining two channels reorder the methods again. Averaged over the four maneuvers, the tremor
that costs SAC and TD3 their bank accuracy also costs them the climb channel, $0.61$--$1.36$~m/s
against $0.10$--$0.33$~m/s for PPO and MPPI. Airspeed goes the other way, and the airframe decides
which method pays for it. On the V7, PPO carries the largest airspeed error of any method,
$3.5$~m/s, as the price of the most accurate learned climb tracking on an airframe with no excess
power; on the Ranger it carries the smallest, $1.1$~m/s, and TD3 the largest, $3.1$~m/s. The same
trade is visible within a single turn in Fig.~\ref{fig:attitude-exec}, where the V7 sinks below its
commanded climb rate to hold $45^\circ$ of bank while the Volantex Ranger loses airspeed instead.
Three caveats belong with these numbers. The LQR has no climb-rate channel and receives
$\dot h^\star$ as an integrated altitude reference, which gives it altitude-error feedback the
learned rate executors lack. MPPI plans over a one-second preview of the stream, so the comparison
is between a planner with lookahead and reactive executors without one. MPPI also picks its plan by
scoring a batch of random candidate control sequences, so each flight is one draw rather than a
fixed control law. Over five draws of an identical cell the V7 circle ranges over
$0.8$--$18.5^\circ$ and the Ranger s-turn over $0.3$--$34.3^\circ$, the widest spread of any cell
in the sweep.


\begin{figure}[t]
\centering
\includegraphics[width=\columnwidth]{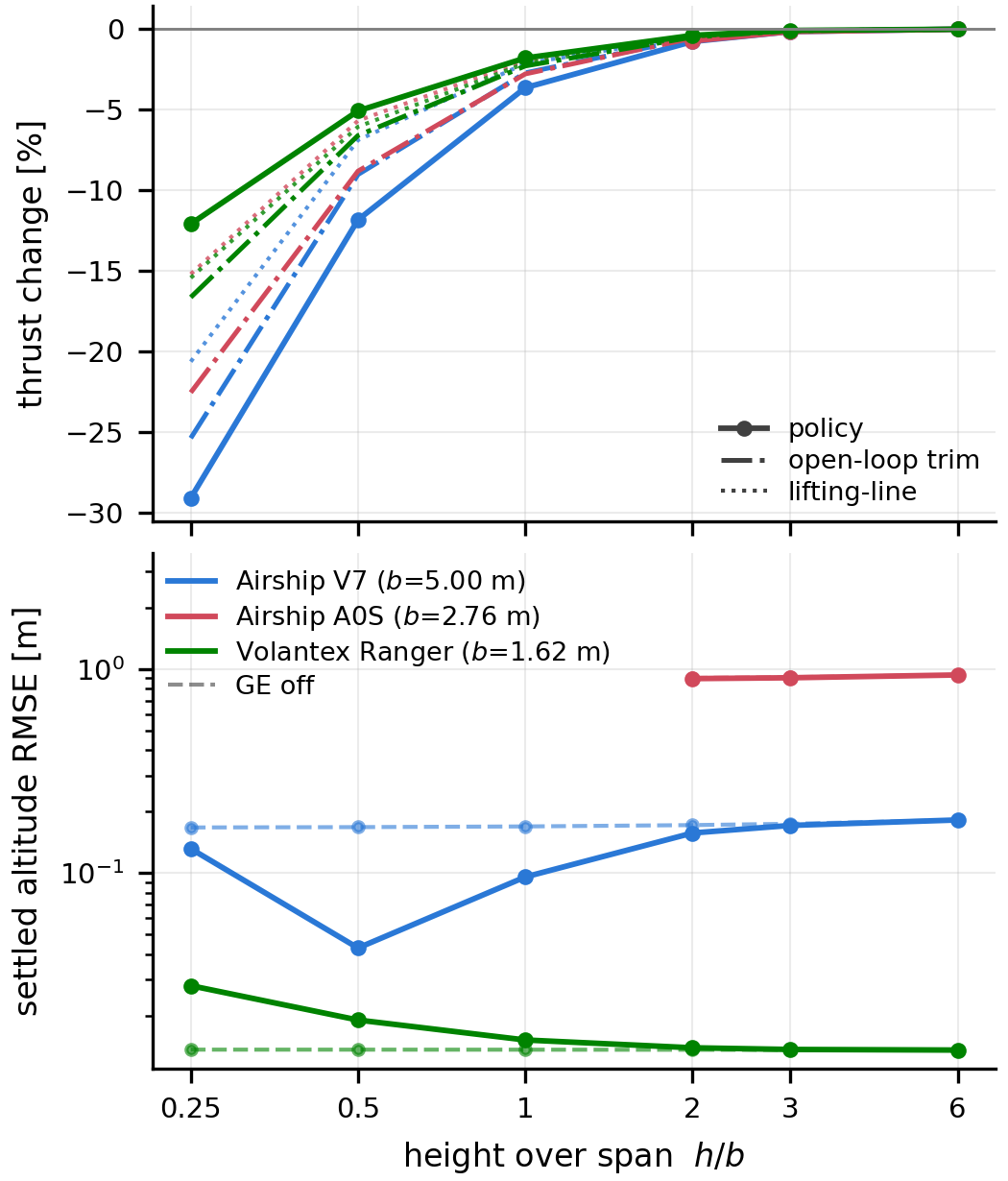}
\caption{\small Ground effect versus $h/b$. Left: thrust change for trim (dash-dot), policy (solid), and theory (dotted). Right: settled altitude error with correction on/off (solid/dashed). Colours denote airframes. A0S has no policy curve below $h/b=2$.}
\label{fig:ge-energy}
\end{figure}

\subsection{Ground Effect and Thrust demand}
\label{sec:ge-energy}

Under a stabilising controller, ground effect changes the thrust required to fly a trajectory rather than the trajectory itself. Table~\ref{tab:ge-energy} and Fig.~\ref{fig:ge-energy} report that change against height over wing span $h/b$, the argument of the lifting-line correction.

Two measurements are given. The first solves a level-flight trim at each height for the angle of
attack, elevator and throttle that null the body forces and the pitching moment, with the correction
on and off. It uses no controller, so it covers every airframe at every band, and it measures what
the airframe offers. The second is the same thrust change measured while a policy holds the band,
and it measures what a controller collects.
At $h/b = 0.25$ the trim thrust falls by \SI{25.4}{\percent} on the Airship~V7,
\SI{22.5}{\percent} on the Airship~A0S and \SI{16.6}{\percent} on the Volantex Ranger, against
predictions of \SI{20.6}{\percent}, \SI{15.2}{\percent} and \SI{15.4}{\percent}. The measurement
exceeds the prediction because the prediction holds the lift coefficient fixed while the aircraft
re-trims: ground effect raises lift, the trim angle of attack falls, and drag falls further. Freezing
the V7 angle of attack at its out-of-ground-effect value of \SI{2.14}{\degree} leaves
\SI{57.8}{\newton} of drag, against \SI{54.4}{\newton} when it is free to re-trim to
\SI{1.68}{\degree}. The closed-loop column follows the same curve without matching it,
\SI{29.1}{\percent} on the V7 and \SI{12.1}{\percent} on the Ranger, because the policy is free to
hold the band at its own airspeed.
Settled tracking error stays below \SI{0.19}{\metre} on the V7 and \SI{0.03}{\metre} on the Ranger
in both arms. The aircraft flies the same trajectory either way, and what ground effect changes is
what that trajectory costs. The A0S rows separate the two questions: the airframe is offered
\SI{22.5}{\percent} at $h/b = 0.25$, and its altitude policy does not hold that band at all.

\begin{table}[t]
\caption{\small Ground-effect impact on thrust. $\Delta T$ is the thrust change; $\Delta T_\pi$ is the policy-controlled change; Theory is the lifting-line prediction.}
\label{tab:ge-energy}
\centering
\setlength{\tabcolsep}{3.5pt}
\begin{tabular}{lrrrrrrr}
\toprule
& $h/b$ & $h$ (m) & $T_{\mathrm{GE}}$ (N) & $T_{\mathrm{no\,GE}}$ (N) & $\Delta T$ & $\Delta T_\pi$ & Theory \\
\midrule
\multirow{6}{*}{\rotatebox{90}{Airship V7}} & 0.25 & 1.25 & 54.39 & 72.89 & $-25.4$\% & $-29.1$\% & $-20.6$\% \\
& 0.50 & 2.50 & 66.32 & 72.89 & $-9.0$\% & $-11.8$\% & $-6.9$\% \\
& 1.00 & 5.00 & 70.92 & 72.89 & $-2.7$\% & $-3.6$\% & $-2.1$\% \\
& 2.00 & 10.00 & 72.49 & 72.89 & $-0.6$\% & $-0.8$\% & $-0.5$\% \\
& 3.00 & 15.00 & 72.81 & 72.89 & $-0.1$\% & $-0.1$\% & $-0.1$\% \\
& 6.00 & 30.00 & 72.90 & 72.90 & $+0.0$\% & $+0.0$\% & $+0.0$\% \\
\midrule
\multirow{6}{*}{\rotatebox{90}{Airship A0S}} & 0.25 & 0.69 & 21.53 & 27.80 & $-22.5$\% & --- & $-15.2$\% \\
& 0.50 & 1.38 & 25.36 & 27.80 & $-8.8$\% & --- & $-5.7$\% \\
& 1.00 & 2.76 & 27.04 & 27.81 & $-2.8$\% & --- & $-1.9$\% \\
& 2.00 & 5.52 & 27.66 & 27.81 & $-0.6$\% & $-0.7$\% & $-0.4$\% \\
& 3.00 & 8.28 & 27.79 & 27.82 & $-0.1$\% & $-0.1$\% & $-0.1$\% \\
& 6.00 & 16.56 & 27.84 & 27.84 & $-0.0$\% & $-0.0$\% & $+0.0$\% \\
\midrule
\multirow{6}{*}{\rotatebox{90}{Volantex Ranger}} & 0.25 & 0.41 & 0.76 & 0.92 & $-16.6$\% & $-12.1$\% & $-15.4$\% \\
& 0.50 & 0.81 & 0.86 & 0.92 & $-6.6$\% & $-5.1$\% & $-6.1$\% \\
& 1.00 & 1.62 & 0.90 & 0.92 & $-2.3$\% & $-1.8$\% & $-2.1$\% \\
& 2.00 & 3.24 & 0.91 & 0.92 & $-0.5$\% & $-0.4$\% & $-0.5$\% \\
& 3.00 & 4.86 & 0.91 & 0.92 & $-0.1$\% & $-0.1$\% & $-0.1$\% \\
& 6.00 & 9.72 & 0.92 & 0.92 & $+0.0$\% & $+0.0$\% & $+0.0$\% \\
\bottomrule
\end{tabular}
\vspace{-0.5cm}
\end{table}


\begin{table}[t]
\caption{\small PPO attitude-executor robustness. Entries are Surv./$\phi$ (bank RMSE, $^\circ$); values use the true simulator state.}
\label{tab:robustness}
\centering
\setlength{\tabcolsep}{2.5pt}
\begin{tabular}{@{}llcccc@{}}
\toprule
Airframe & Disturbance & Nom. & Mild & Med. & Sev. \\
\midrule
\multirow{4}{*}{Airship V7}
& Sensor noise  & 1.00/2.0 & 1.00/2.0 & 1.00/2.0 & 1.00/2.0 \\
& Sensor delay  & 1.00/2.0 & 1.00/7.0 & 1.00/12.0 & 1.00/20.6 \\
& Action delay  & 1.00/2.0 & 1.00/7.2 & 1.00/14.5 & 1.00/21.7 \\
& Turbulence    & 1.00/2.0 & 1.00/7.7 & 0.75/18.7 & 1.00/22.0 \\
\midrule
\multirow{4}{*}{Volantex Ranger}
& Sensor noise  & 1.00/1.3 & 1.00/1.3 & 1.00/1.3 & 1.00/1.4 \\
& Sensor delay  & 1.00/1.3 & 1.00/1.3 & 1.00/1.3 & 1.00/9.7 \\
& Action delay  & 1.00/1.3 & 1.00/1.3 & 1.00/1.3 & 1.00/11.7 \\
& Turbulence    & 1.00/1.3 & 1.00/1.9 & 1.00/5.5 & 0.00/--- \\
\bottomrule
\end{tabular}
\end{table}

\begin{figure}[t]
\centering
\includegraphics[width=\columnwidth]{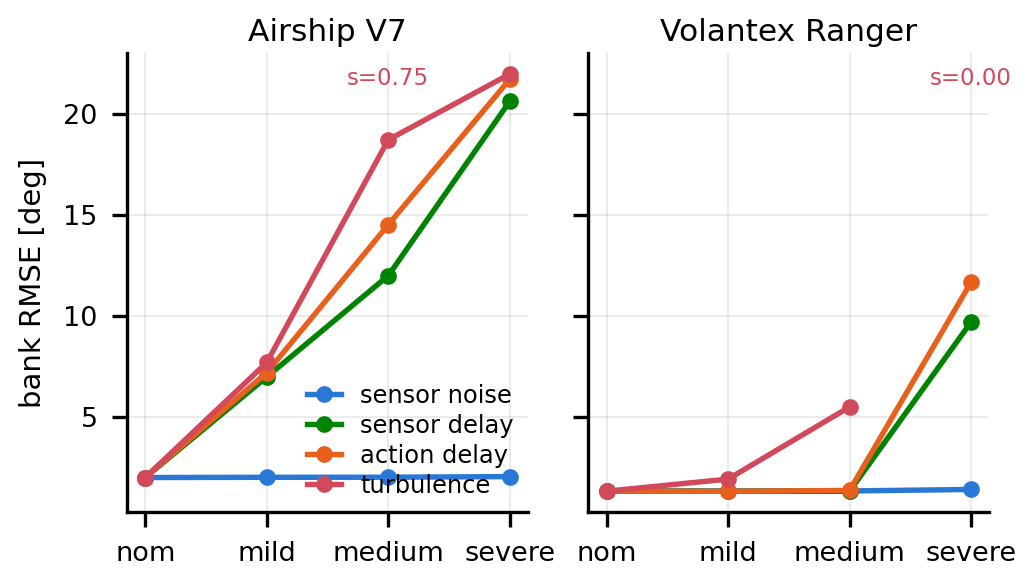}
\caption{\small Bank RMSE of the PPO executor against disturbance severity, one line per family, for the
two airframes. Survival is annotated only where it leaves 1.00.
}
\label{fig:robustness}
\end{figure}

\subsection{Robustness to Sensing, Latency and Turbulence}
\label{sec:robustness}

The maneuvers of Section~\ref{sec:maneuvers} assume a perfect sensor, an instantaneous command
path and still air. Table~\ref{tab:robustness} and Fig.~\ref{fig:robustness} relax each of those in
turn on the PPO executor, over four disturbance families at three severities set in physical units:
per-channel sensor noise floors, \SIrange{20}{80}{\milli\second} of observation latency,
\SIrange{20}{100}{\milli\second} of command latency, and a Dryden field at
$\sigma_u/V_a = 0.05$ to $0.15$. A disturbance corrupts what the policy sees or commands; every
metric is computed from the true simulator state.

The families do not cost the same. Sensor noise is free: bank RMSE holds at \SI{2.0}{\degree} on the
Airship~V7 and moves from \SI{1.3}{\degree} to \SI{1.4}{\degree} on the Volantex Ranger across the
whole grid, because the executor reacts to a state it can average rather than to a single sample.
Latency is expensive, and the two paths cost the same: \SI{20.6}{\degree} and \SI{21.7}{\degree} on
the V7 at the severe grid, an order of magnitude above nominal. This is the classical dead-time
penalty on an attitude loop. Turbulence is the only family that ends a run. The V7 loses one
maneuver of four at $\sigma_u/V_a = 0.10$ and the Ranger departs on all four at $0.15$. Each wind
cell is a single gust realisation, so the V7 recovering to full survival at $0.15$ is a draw of the
field and not a trend.

The airframe matters as much as the disturbance, and not in the direction the nominal table
suggests. The Ranger absorbs both latencies to the medium grid with no measurable loss,
\SI{1.3}{\degree} throughout, while the V7 degrades from the mild grid onward. Yet it is the Ranger
that departs first in wind. Reporting a single disturbance magnitude would have collapsed these
three orderings into one number.


\section{Conclusions}
This paper presented a fixed-wing simulation framework for fixed-wing flight control in which learned and classical controllers can
drive different airframe models in a variety of flight regimes. Defining tasks as command
streams rather than waypoints evaluates controllers on the states they must continuously maintain.
Learned methods are scored over
three training seeds on both tracking tasks and reported with their spread. Ground effect is
reported as the thrust required to fly, separating what measured GE from a free flying airframe to a policy
driven flight, and robustness by disturbance severity over four different parameters.
No method orders consistently: MPPI is the most accurate attitude executor where it converges yet
acquires neither commanded altitude, the LQR is the most precise regulator yet acquires half the
episodes on one airframe, and PPO is the most consistent method. 
Future work adds a guidance layer above the attitude executor for sim-to-real transfer and consider the ground-effect
to fly over water surfaces with waves.


\small
\bibliographystyle{IEEEtran}
\bibliography{references}

\end{document}